\documentclass{article}

\usepackage[dblblindworkshop, final]{neurips_2026}

\usepackage{subcaption}
\usepackage[utf8]{inputenc} %
\usepackage[T1]{fontenc}    %
\usepackage{hyperref}       %
\usepackage{inconsolata}
\usepackage{url}            %
\usepackage{booktabs}       %
\usepackage{amsmath}
\usepackage{amsfonts}       %
\usepackage{amssymb}
\usepackage{amsthm}
\usepackage{nicefrac}       %
\usepackage{microtype}      %
\usepackage{xcolor}         %

\usepackage{graphicx}
\usepackage{listings}
\lstdefinestyle{problog}{
    basicstyle=\ttfamily\footnotesize,
    commentstyle=\itshape\color{gray},
    morecomment=[l]{\%},
    breaklines=true, columns=fullflexible, keepspaces=true,
    frame=single, framesep=4pt, xleftmargin=6pt,
}
\usepackage[section,below]{placeins}

\usepackage{enumitem}

\newtheorem{theorem}{Theorem}

\newtheorem{lemma}{Lemma}
\newtheorem{assumption}{Assumption}

\usepackage{xcolor}

\title{Safe Reinforcement Learning via \\ Probabilistic Option Shielding}
\workshoptitle{Neuro-Symbolic Embodied Intelligence}

\author{%
  Gian Luca Ascari\thanks{Work done while at KU Leuven.}\\
  Apixa\\
  \texttt{gianlu.ascari@gmail.com}
  \And
  Isidoro Tamassia \\
  KU Leuven \\
  \texttt{isidoro.tamassia@kuleuven.be}
  \And
  Ying Jiao \\
  KU Leuven \\
  \texttt{ying.jiao@kuleuven.be}
  \And
  Luc De Raedt \\
  KU Leuven, Örebro University \\
  \texttt{luc.deraedt@kuleuven.be}
  \And
  Giuseppe Marra \\
  KU Leuven \\
  \texttt{giuseppe.marra@kuleuven.be}
}

\begin{document}

\maketitle

\begin{abstract}

Shielding is a safe reinforcement learning technique that provides safety guarantees while learning, by blocking unsafe actions. Traditional shielding relies on a complete model of the environment to provide long-term guarantees. Probabilistic logic shielding (PLS) lifts the requirement of having such model, but only considers safety in the short term, typically looking only one or few steps ahead. We propose Probabilistic Option Shielding (POS), which extends PLS to the option level, recovering long-horizon safety without reintroducing the need for a complete environment model. POS learns a model-free critic directly from one-step PLS signals, to estimate the probability that an entire option will complete safely under uncertainty. We evaluate POS on safe navigation tasks in the LogiCity environment, and show that it achieves considerably safer training while matching the task performance of standard step-level PLS and other safety related reinforcement learning baselines.

\end{abstract}

\section{Introduction} 
The safety of an agent acting in an environment typically depends not only on the immediate effects of its actions but also on their future consequences. For example, accelerating towards a green traffic light while driving a car may be safe in the moment, but later leave too little distance to stop safely once it becomes red. In reinforcement learning (RL), such consequences can be addressed with options: temporally extended behaviors composed of multiple actions. Our goal is to propose an effective way of enforcing long-term behavioral safety by combining options with probabilistic logic shielding methods. 

Shielding provides a principled approach to safety in RL by evaluating candidate actions based on safety knowledge. Classical approaches, rooted in model checking and formal verification~\citep{baier2008principles}, assume precise safety specifications and rely on a model of the environment to enforce them throughout execution \citep{alshiekh2018shielding}. Subsequent work has extended shielding to allow for uncertainty in safety specifications \citep{carr2023partial, jansen2020safe, yang2023safe, belardinelli2025probabilistic}. In particular, Probabilistic Logic Shields (PLS) \citep{yang2023safe} avoid the need for a transition model by encoding safety conditions directly in a probabilistic (logic) program. The program computes action-safety probabilities, which are used to reweight the agent's policy while maintaining the overall process differentiable, and can therefore be integrated into state-of-the-art policy-gradient methods \citep{schulman2015trust, schulman2017proximal}. However, approaches like PLS do not directly consider the long-term consequences that an action may cause. Although multi-step safety could, in principle, be captured by modeling future transitions within the probabilistic logic program, doing so would reintroduce the need for knowledge of the environment dynamics and substantially increase the cost of inference.

The options framework~\citep{sutton1999between}, employed by most approaches to Hierarchical RL (HRL, \citep{pateria2021hierarchical}), provides a natural level of abstraction for addressing
these limitations.
Rather than choosing a new low-level action at every
step, a high-level policy selects an option and the agent commits to its execution until a termination condition triggers. Shielding at this level requires assigning each option the probability that captures whether its entire execution satisfies the safety specification. Unfortunately, computing this probability on-the-fly also generally requires knowledge of the transition model. 

In this work, we show that the safety of an option can be learned without access to a model of the environment dynamics. More specifically, we propose Probabilistic Option Shielding (POS), which learns an option-safety critic by temporal-difference (TD) updates from observed transitions~\citep{sutton2018reinforcement}, either in a pretraining phase or together with the policy. The learned critic is then used to reweigh the option-policy analogously to the PLS action-level shielding, thereby reducing the probability of unsafe behavior.

\paragraph{Contributions.}
\begin{itemize}
    \item We lift PLS to the options framework, extending the short-sighted traditional PLS to long-term behaviors. 
    \item We show that under mild assumptions, 
the resulting option-safety operator has a unique fixed point and can be optimized efficiently.
    \item We estimate the safety of options by TD learning with a safety critic, realized either tabularly or as a neural network, and either pretrained or updated online during policy learning. We demonstrate experimentally that this yields considerably safer training at comparable returns to the baselines.
    \item We release an open-source implementation.\footnote{\url{https://github.com/gianlucaascari/probabilistic-option-shielding.git}}
\end{itemize}

\section{Background}
\label{sec:background}
\subsection{Markov Decision Processes}
\label{sec:background_mdp}

We consider a discounted Markov decision process (MDP,~\cite{puterman2014markov}) $\mathcal{M}=(\mathcal{S},\mathcal{A},P,R,\rho_0,\gamma),$ where $\mathcal{S}$ is the state space, $\mathcal{A}$ is the set of
(low-level) actions, $P(s' \mid s,a)$ is the transition probability,
$R(s,a,s')$ is the expected one-step reward, $\rho_0$ is the initial-state
distribution, and $\gamma \in [0,1)$ is the discount factor. %
At time step $t$, the agent observes a state $s_t$, samples an
action $a_t \sim \pi(\cdot \mid s_t)$, and the environment samples
a successor state according to $s_{t+1} \sim P(\cdot \mid s_t,a_t).$
The interaction therefore produces a trajectory
$s_0,a_0,r_0,s_1,a_1,r_1,\ldots$, where
$s_0 \sim \rho_0$ and
$\mathbb{E}[r_t \mid s_t,a_t,s_{t+1}]
=R(s_t,a_t,s_{t+1})$.
The objective is to find a policy that maximises the expected discounted
return
\[
    J(\pi)
    =
    \mathbb{E}_{\pi,P}
    \left[
        \sum_{t=0}^{\infty} \gamma^t r_t
    \right].
\]

\subsection{Options and Temporal Abstraction}
\label{sec:background_options}

The options framework extends an MDP with temporally extended actions
\citep{sutton1999between}. A Markov option is a tuple $o = \left(\mathcal{I}_o,\pi_o,\beta_o\right)$ where $\mathcal{I}_o \subseteq \mathcal{S}$ is the initiation set,
$\pi_o(a \mid s)$ is the intra-option policy over low-level actions, and
$\beta_o(s) \in [0,1]$ is the probability that the option terminates upon
reaching state $s$. The set of options available in state $s$ is $\mathcal{O}(s)
    =
    \{o \in \mathcal{O} : s \in \mathcal{I}_o\}.$
When the agent
selects option $o$ in state $s_t$, the option repeatedly samples primitive
actions according to
$a_t \sim \pi_o(\cdot \mid s_t)$. After arriving in $s_{t+1}$, the option terminates with probability
$\beta_o(s_{t+1})$; otherwise, control remains with the same option.
Consequently, an option executes for a random duration $T_o \geq 1$ and
produces the state-action sequence $s_0,a_0,s_1,\ldots,a_{T_o-1},s_{T_o}.$ An option also terminates if the underlying episode ends.

A high-level policy over options $\Pi(o \mid s),\, o \in \mathcal{O}(s)$
selects an option whenever the previous one terminates. The selected
option then retains control until its own termination condition is
satisfied. Thus, decisions made by $\Pi$ may occur at irregular time intervals. The resulting high-level decision process is also called a 
semi-Markov decision process (SMDP).

\subsection{Probabilistic Logic Shields}
\label{sec:background_pls}

Shielding incorporates prior safety knowledge into an agent's decisions by
preventing actions that violate a given specification
\citep{alshiekh2018shielding}. Probabilistic Logic Shields (PLS) extend this
idea to specifications involving uncertain facts
\citep{yang2023safe}. In PLS, a probabilistic logic program (ProbLog, \cite{de2007problog}) encodes the
safety specification and, given a state $s$ and candidate primitive action
$a$, computes $\mathbf P(\mathrm{safe} \mid s,a) \in [0,1].$
This quantity represents the probability, according to the encoded
specification and a safety-relevant description of the state $s$, that executing $a$ in $s$ is
safe. The probability of each base fact may be provided by a neural network. 

For an action policy $\pi$, the probability of selecting a safe action in
state $s$ is obtained by marginalizing over the policy:
\[
    \mathbf P_{\pi}(\mathrm{safe} \mid s)
    =
    \sum_{a \in \mathcal{A}}
        \mathbf P(\mathrm{safe} \mid s,a)\pi(a \mid s).
\]
PLS constructs a shielded policy by conditioning the original policy on
safety:
\[
    \pi^{+}(a \mid s)
    =
    \frac{
        \mathbf P(\mathrm{safe} \mid s,a)\pi(a \mid s)
    }{
        \mathbf P_{\pi}(\mathrm{safe} \mid s)
    },
    \label{eq:pls_policy}
\]
Thus, actions retain their
relative probability under $\pi$ but are also weighted according to their
probability of satisfying the safety specification. Because ProbLog programs are differentiable~\citep{de2007problog}, the shielded policy can be incorporated into any
policy-gradient method~\citep{williams1992simple, sutton1999policy}, including popular neural methods such as TRPO~\citep{schulman2015trust} and PPO~\citep{schulman2017proximal}, while retaining the underlying learning objective.

As described, PLS evaluates the safety of the next low-level
action. Looking multiple steps ahead is possible when future transitions
are represented inside the probabilistic logic program, but this may be cumbersome to specify exactly. 
Moreover, performing multi-step planning increases sharply the cost of inference. These limitations motivate us to apply shielding to
temporally extended decisions compactly represented by options.

\section{Methodology}
\label{sec:method}
We now illustrate our method for temporally extended probabilistic shielding, which we call Probabilistic Option Shielding (POS). POS combines concepts from the neurosymbolic PLS method and the option framework detailed above.  
We begin by defining the probability of option-level safety and show that it admits a recursive computation which provably converges to a fixed point. Then, we detail how we can learn a safety critic that estimates this quantity in practical settings with temporal-difference (TD) learning techniques~\citep{sutton2018reinforcement} from one-step safety signals. Note that we assume the option set is given; extending our method by also learning a meaningful set of options from low-level actions is left for future work. 

\subsection{Option Safety}
\label{sec:method:def}

An option $o$ executes for a random number of steps $T_o \geq 1$. We define its cumulative safety as the expected product of the per-step safety probabilities along its execution:
\begin{equation}
    \mathbf{P}(\mathrm{safe}\mid s,o)
    =
    \mathbb{E}_{\substack{
        \tau\sim(P,\pi_o,\beta_o,
        s_0=s)
    }}
    \left[
        \prod_{t=0}^{T_o-1}
        \mathbf{P}(\mathrm{safe}\mid s_t,a_t)
    \right],
    \label{eq:option-safety-def}
\end{equation}
where $\tau=(s_0,a_0,\ldots,s_{T_o})$ is the trajectory determined by the intra-option policy $\pi_o$, the environment dynamics $P$, and the option-termination probability $\beta_o$. Assuming that the per-step safety events are conditionally independent given the state-action trajectory, the product inside the expectation is the probability that every primitive action along a particular execution of $o$ is safe. The expectation then averages this probability over all possible executions of the option from $s$.

The high-level shielded policy is obtained by applying the PLS reweighting rule to options:
\begin{equation}
    \Pi^+(o\mid s)
    =
    \frac{
        \mathbf{P}(\mathrm{safe}\mid s,o)\Pi(o\mid s)
    }{
        \mathbf{P}_{\Pi}(\mathrm{safe}\mid s)
    },
    \qquad
    \mathbf{P}_{\Pi}(\mathrm{safe}\mid s)
    =
    \sum_{o\in\mathcal{O}(s)}
    \mathbf{P}(\mathrm{safe}\mid s,o)\Pi(o\mid s).
    \label{eq:shielded-option-policy}
\end{equation}

A primitive action $a$ can be represented as a one-step option $o_a$ whose intra-option policy selects $a$ deterministically. In this case,
\[
    \mathbf{P}(\mathrm{safe}\mid s,o_a)
    =
    \mathbf{P}(\mathrm{safe}\mid s,a),
\]
so the proposed formulation recovers action-level shielding as a special case. The remaining challenge is to estimate $\mathbf{P}(\mathrm{safe}\mid s,o)$ without manually specifying the safety of every option.

\subsection{Recursive Form and Iterative Convergence}
\label{sec:method:bellman}

Equation~\eqref{eq:option-safety-def} admits a recursive decomposition. Conditioning on the first action $a$ and the resulting state $s'$, the option either terminates at $s'$, in which case no further safety terms are accumulated, or continues from $s'$. Combining these cases using the termination probability $\beta_o(s')$ gives
\begin{equation}
    \mathbf{P}(\mathrm{safe}\mid s,o)
    =
    \mathbb{E}_{\substack{
        a\sim\pi_o(\cdot\mid s)\\[-0.5mm]
        s'\sim P(\cdot\mid s,a)
    }}
    \left[
        \mathbf{P}(\mathrm{safe}\mid s,a)
        \left(
            \beta_o(s')
            +
            \bigl(1-\beta_o(s')\bigr)
            \mathbf{P}(\mathrm{safe}\mid s',o)
        \right)
    \right].
    \label{eq:bellman}
\end{equation}
The continuation probability $1-\beta_o(s')$ acts as a state-dependent discount. If the option terminates at $s'$, the current action-safety probability is the only contribution. If it continues, this probability is multiplied by the cumulative safety of executing the remainder of the option from $s'$.

We now show that repeatedly applying Equation~\eqref{eq:bellman} converges to a fixed point. This establishes that $\mathbf{P}(\mathrm{safe}\mid s,o)$ is well defined and can be computed by dynamic programming or estimated using methods similar to value-function evaluation \citep{bellman1966dynamic}. Let $\mathcal{S}$ be the finite set of states reachable under the option set, and define the safety operator $\mathcal{T}^o$ on $[0,1]^{\mathcal{S}}$ by
\begin{equation}
    (\mathcal{T}^o f)(s)
    =
    \mathbb{E}_{\substack{
        a\sim\pi_o(\cdot\mid s)\\[-0.5mm]
        s'\sim P(\cdot\mid s,a)
    }}
    \left[
        \mathbf{P}(\mathrm{safe}\mid s,a)
        \left(
            \beta_o(s')
            +
            \bigl(1-\beta_o(s')\bigr)f(s')
        \right)
    \right].
    \label{eq:safety-operator}
\end{equation}

\begin{assumption}[Bounded option duration]
\label{ass:bounded}
There exists a finite $N$ such that, for every $s\in\mathcal{S}$,
\[
    \mathbf{P}(T_o\leq N\mid s_0=s)=1,
\]
where $T_o\geq 1$ is the number of steps executed by option $o$ before termination.
\end{assumption}

Assumption~\ref{ass:bounded} requires options to terminate within a uniformly bounded number of steps. It is satisfied, for example, when options have an explicit maximum duration or when they are executed within finite-horizon episodes. Under this assumption, Equation~\eqref{eq:bellman} is the fixed-point equation $\mathcal{T}^o f=f$. The proof of the following result is given in Appendix~\ref{app:fixed-point}.

\begin{theorem}[Fixed point]
\label{thm:fixed-point}
Under Assumption~\ref{ass:bounded}, $\mathcal{T}^o$ admits a unique fixed point in $[0,1]^{\mathcal{S}}$. Moreover, for any initial $f\in[0,1]^{\mathcal{S}}$, the sequence
\[
    f,\ \mathcal{T}^o f,\ (\mathcal{T}^o)^2f,\ \ldots
\]
reaches this fixed point after at most $N$ applications of $\mathcal{T}^o$.
\end{theorem}

\subsection{Temporal-Difference Safety Evaluation}
\label{sec:method:td}

Evaluating the expectation in Equation~\eqref{eq:bellman} analytically requires the transition model $P(s'\mid s,a)$, which is unavailable in the model-free setting. We therefore estimate it using temporal-difference (TD) learning \citep{sutton2018reinforcement}. We maintain a \emph{safety critic} $\hat{P}(\mathrm{safe}\mid s,o)$, initialized in $[0,1]$ for each state-option pair, and update it from observed transitions.

For a transition $(s_t,a_t,s_{t+1})$ generated during the execution of option $o$, the bootstrapped TD target is
\begin{equation}
    y_t
    =
    \mathbf{P}(\mathrm{safe}\mid s_t,a_t)
    \left[
        \beta_o(s_{t+1})
        +
        \bigl(1-\beta_o(s_{t+1})\bigr)
        \hat{P}(\mathrm{safe}\mid s_{t+1},o)
    \right].
    \label{eq:td-target}
\end{equation}
The critic estimate at $s_t$ is updated with step size $\alpha_n\in[0,1]$:
\begin{equation}
    \hat{P}(\mathrm{safe}\mid s_t,o)
    \leftarrow
    \hat{P}(\mathrm{safe}\mid s_t,o)
    +
    \alpha_n
    \left[
        y_t-\hat{P}(\mathrm{safe}\mid s_t,o)
    \right].
    \label{eq:td-update}
\end{equation}

The immediate term $\mathbf{P}(\mathrm{safe}\mid s_t,a_t)$ is already computed by the PLS shield, so evaluating the target requires no additional probabilistic-logic inference. Because the term in brackets in Equation~\eqref{eq:td-target} is a convex combination of $1$ and $\hat{P}(\mathrm{safe}\mid s_{t+1},o)$, the target $y_t$ lies in $[0,1]$. Equation~\eqref{eq:td-update} is itself a convex combination of the current estimate and the target. Consequently, the critic remains in $[0,1]$ throughout learning, provided that it is initialized in this interval. 

When the state space is sufficiently small, the safety critic can be represented as a table with one entry $\hat{P}(\text{safe}\mid s,o)$ for each state-option pair $(s,o)$. After observing a transition generated by executing option $o$, the corresponding entry is updated toward the TD target using Eq.~\eqref{eq:td-update}. For larger or continuous state spaces, we can instead approximate the critic with a neural network $\hat{P}_\theta(\text{safe}\mid s,o)$. Its parameters are trained by semi-gradient TD learning~\citep{sutton2018reinforcement}, minimizing the squared error between the current estimate and the bootstrapped target while treating the latter as constant during each update. 

Importantly, the option-safety fixed point does not depend on the option-level policy $\Pi$: changing $\Pi$ only affects which state-option pairs are visited. Consequently, transitions collected under any option-level policy can be used to pretrain the critic. When a separate data-collection phase is impractical, the critic can be learned online using the transitions collected during task-policy learning.
We evaluate both the pretrained and online regimes.

\begin{figure}[t]
    \centering

    \begin{subfigure}[t]{0.48\linewidth}
        \centering
        \includegraphics[width=0.47\linewidth]{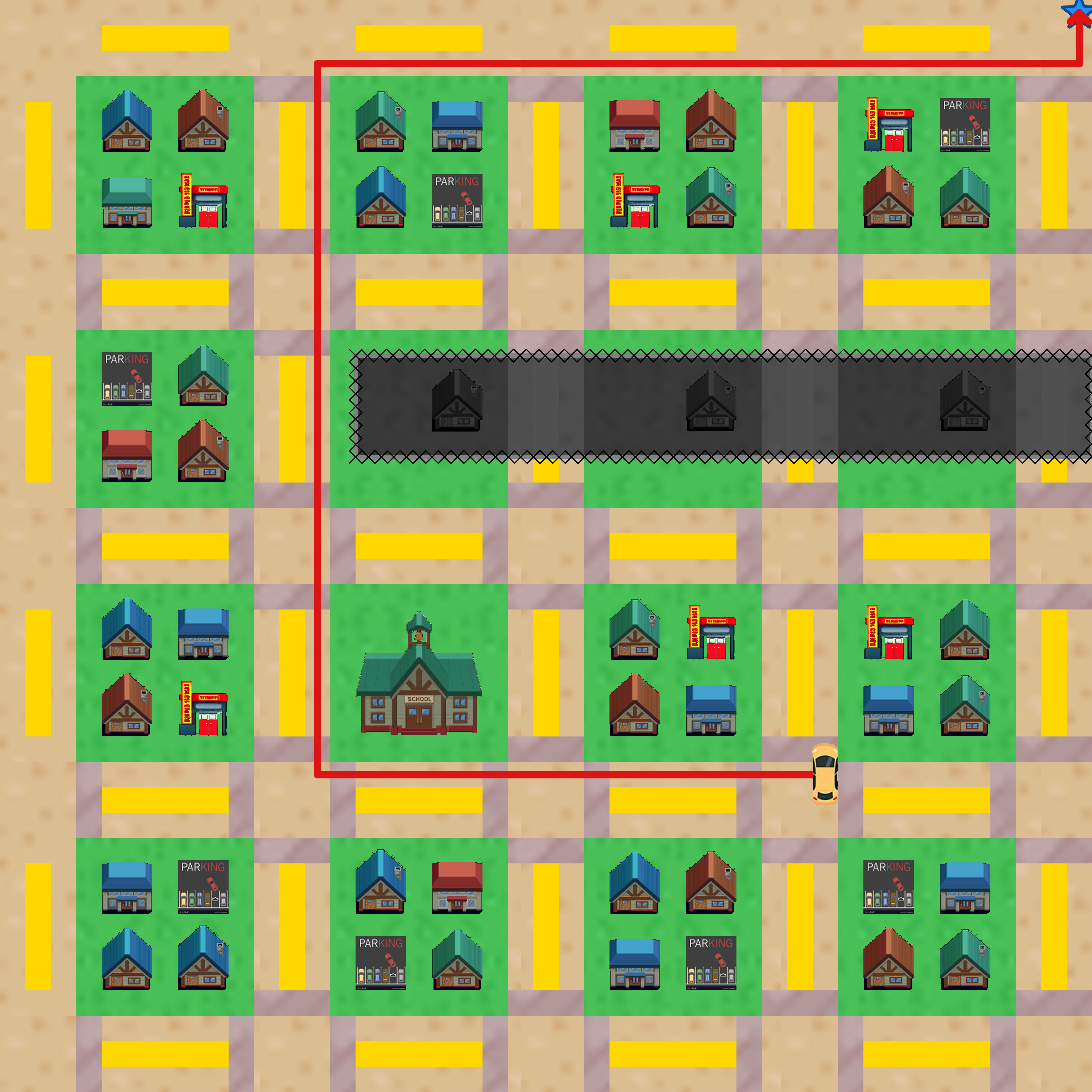}\hfill
        \includegraphics[width=0.47\linewidth]{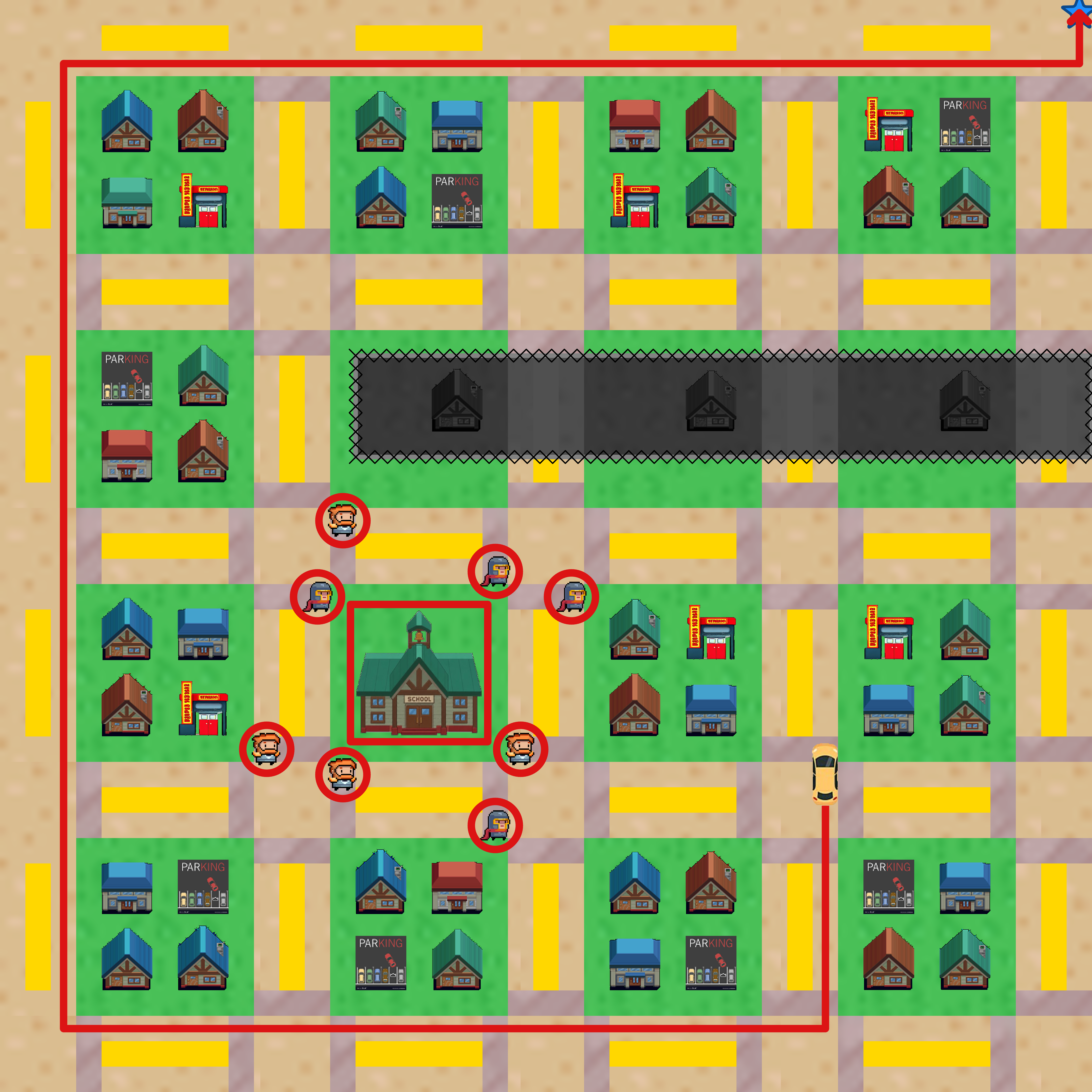}
        \caption{\textbf{Large Map.} \texttt{school-time} off (L)
        and on (R).}
        \label{fig:large-map}
    \end{subfigure}
    \hfill
    \begin{subfigure}[t]{0.48\linewidth}
        \centering
        \includegraphics[width=0.47\linewidth]{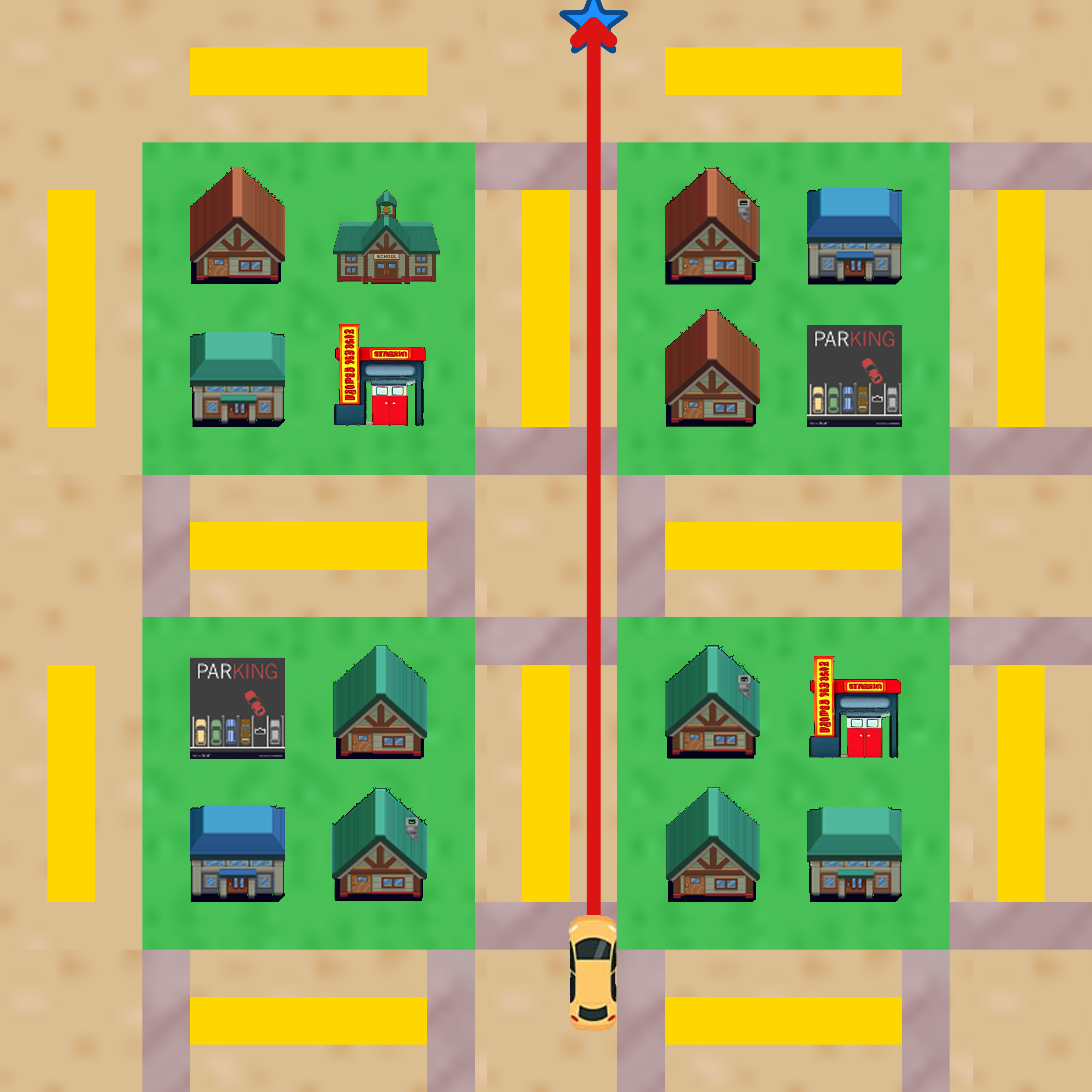}\hfill
        \includegraphics[width=0.47\linewidth]{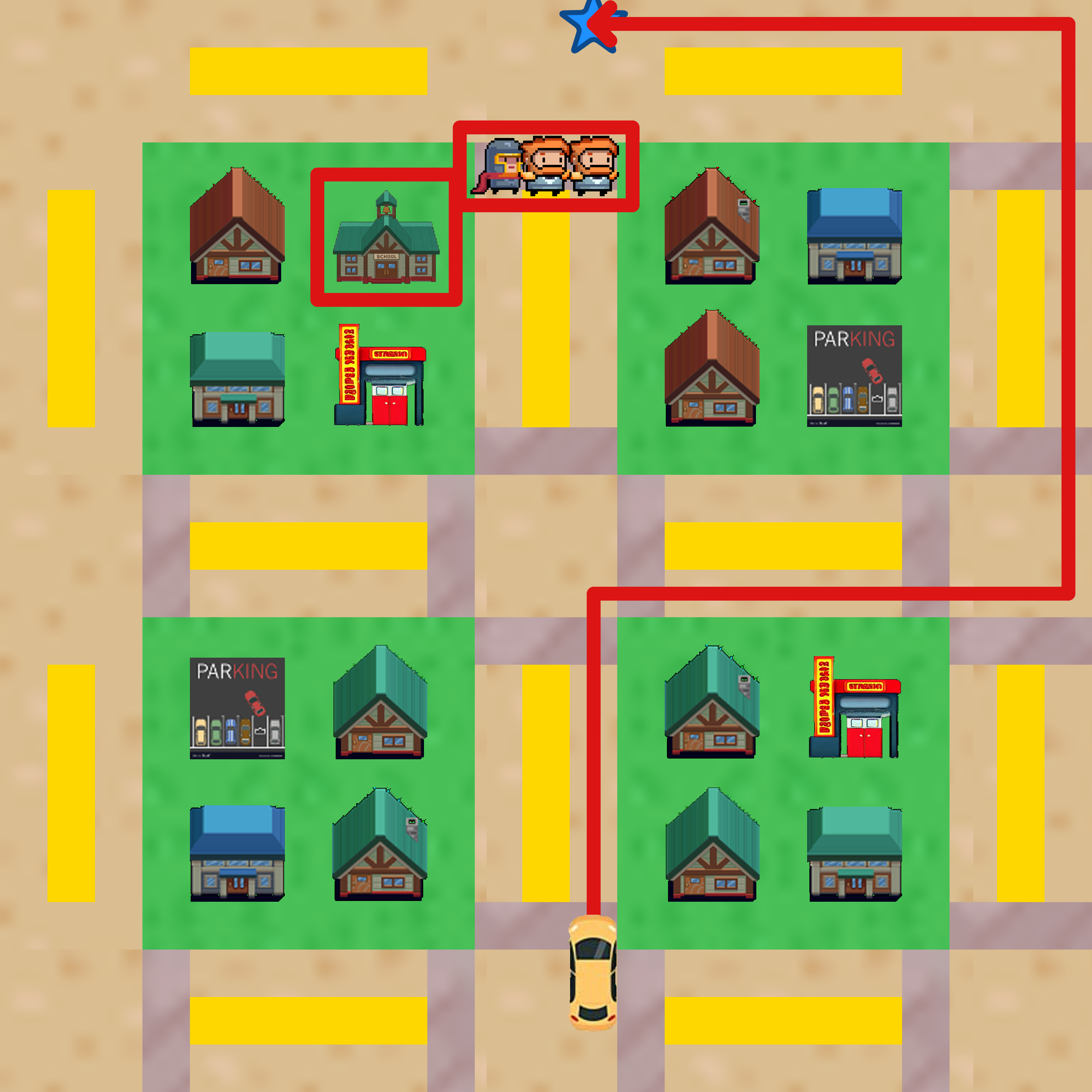}
        \caption{\textbf{Stochastic Map.} \texttt{school-time} off
        (L) and on (R).}
        \label{fig:stochastic-map}
    \end{subfigure}

    \caption{Visualizations of the \emph{Large Map} and the \emph{Stochastic Map} and the corresponding safety concerns with \texttt{school-time} off or on. Specifically, the black area indicates a military zone, while pedestrians are circled in red. The red arrow is the optimal safe trajectory from the start to the goal.}
    \label{fig:maps}
\end{figure}

\section{Experiments}
\label{sec:experiments}

\subsection{Setup}
\label{sec:exp:setup}

\paragraph{Environment.}

We evaluate our approach in LogiCity~\citep{li2024logicity}, a recently proposed neurosymbolic AI benchmark.
In each episode, the agent starts at a fixed grid cell and must reach a
fixed goal by selecting one of five primitive actions:
\emph{left}, \emph{right}, \emph{up}, \emph{down}, or \emph{stop}.
At each step, the agent observes its current coordinates $(x,y)$ and a
binary \texttt{school-time} variable. The latter is randomly sampled as
$\operatorname{Bernoulli}(0.5)$ at the beginning of each episode and
remains fixed throughout the episode. When \texttt{school-time} is
active, pedestrians occupy cells along the shortest path to the goal,
making that route unsafe. An episode terminates when the agent reaches
the goal, violates a safety constraint, or reaches a fixed step limit. The
agent receives a reward of $1$ for reaching the goal and $0$
otherwise.

We consider two variants, the \emph{Large Map}
(\autoref{fig:large-map}) and the \emph{Stochastic Map} (\autoref{fig:stochastic-map}). The \emph{Large Map}
includes pedestrians who only show up during \texttt{school-time}, as well as a military zone that is always unsafe (independently of
\texttt{school-time}). We use this map to evaluate whether the
proposed method can enforce both static and context-dependent safety
constraints over longer horizons: as shown in \autoref{fig:large-map}, the optimal safe path requires a long detour. The \emph{Stochastic Map} is smaller but
introduces a simulated \emph{inertia} effect: at each step, with probability $0.3$, the
agent is forced to repeat the previously executed low-level action rather
than the one it would have selected. Intuitively, this requires the agent to be more far-sighted since trying to brake or change direction too close to the pedestrians may fail due to the inertia.

\paragraph{Method.} Safety is specified by a ProbLog program~\citep{de2007problog} (see Appendix~\ref{app:spec}) that maps the four adjacent-cell sensors and the selected action to $\mathbf{P}(\text{safe}\mid s, a)$, as in the standard PLS setup~\citep{yang2023safe}.
We take the rules to be correct and the sensor readings to be noise-free. Then, four options are defined, one per cardinal direction. Each navigates from the agent's current cell to the next intersection in that direction by executing the corresponding low-level actions and terminates on arrival, or earlier if the episode ends. The safety of each option is estimated via a safety critic, which is instantiated in a neural form (a comparison with a tabular instantiation is available in Appendix~\ref{app:neur-vs-tab}). The resulting estimates are used to shield the options as described in \autoref{sec:method}. As the underlying policy-learning method, we choose PPO \citep{schulman2017proximal}, a state-of-the art model-free RL algorithm. 
We test our method (\texttt{POS-PPO}) both when the critic is pretrained and when it is learned online alongside the policy. Further details, including the safety program, the option definition, and the critic instantiation can be found in Appendix~\ref{app:hyper}.

\paragraph{Baselines and metrics.}
We compare our approach to five distinct baselines: unshielded \texttt{PPO} over low-level actions~\citep{schulman2017proximal}, PPO with standard one-step PLS shielding (\texttt{PPO-PLS}, \cite{yang2023safe}), unshielded PPO over options (\texttt{OptionPPO}), PPO over options with additional reward-shaping (\texttt{OptionPPO-RS}), and PPO over options with the safety constraints modeled as in a constrained MDP~\citep{altman1999constrained} and enforced with a Lagrangian approach (\texttt{OptionPPO-Lag},~\citealp{tessler2019reward, ray2019benchmarking}). Details on the implementations and hyperparameter tuning of each method can be found in Appendix~\ref{app:hyper}. For each experiment, we report cumulative safety violations and episodic discounted return averaged over 10 seeds with the corresponding standard error. 

\paragraph{Experimental questions.} Our experiments aim to address the following questions:

\begin{enumerate}
  \item[\textbf{Q1.}] How accurate are the safety estimates learned by our tabular and neural option-safety critics?
  \item[\textbf{Q2.}] Does \texttt{POS-PPO} yield safer and comparably rewarding policies to baseline approaches?
\end{enumerate}

\begin{figure}[t]
    \centering

    \begin{subfigure}[t]{0.49\linewidth}
        \includegraphics[width=\linewidth]{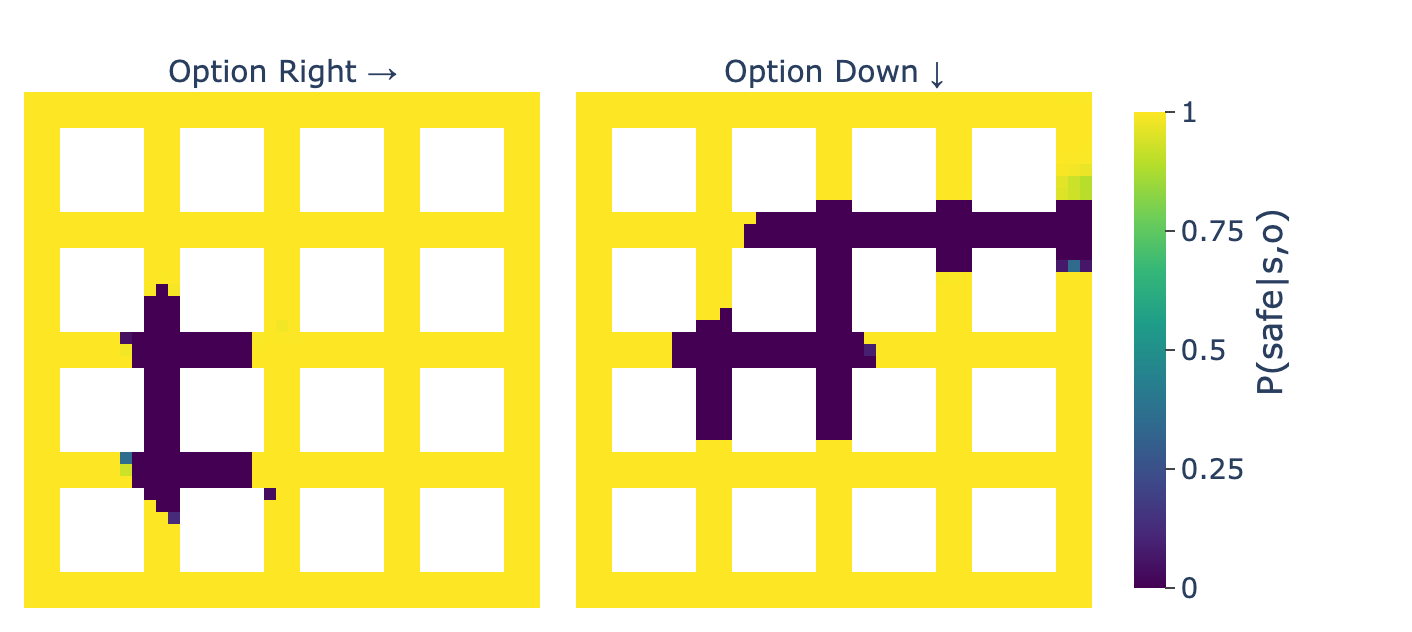}
        \caption{\textbf{Large map.} Options \texttt{right} (L) and \texttt{down} (R).}
        \label{fig:large-map-heatmap}
    \end{subfigure}
    \hfill
    \begin{subfigure}[t]{0.48\linewidth}
        \centering
        \includegraphics[width=\linewidth]{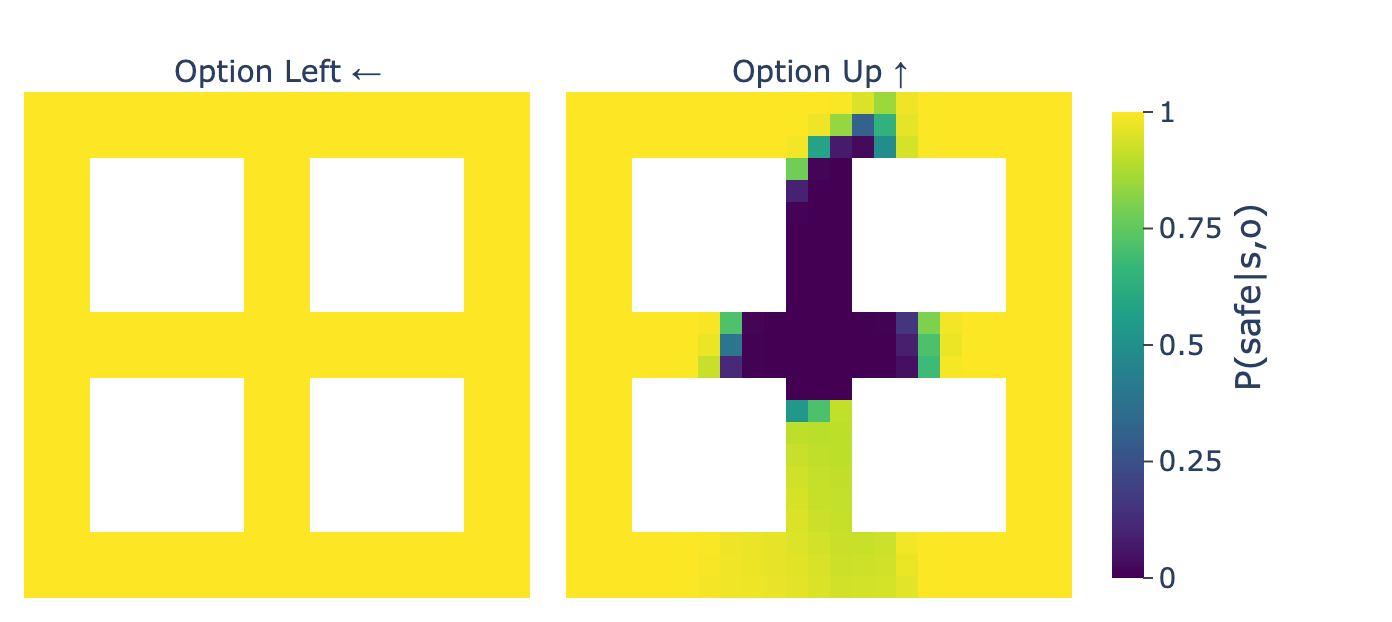}
        \caption{\textbf{Stochastic map.} Options \texttt{left} (L) and \texttt{up} (R).}
        \label{fig:inertia-map-heatmap}
    \end{subfigure}

    \caption{\textbf{Learned option safety.} Examples neural critic estimates for \emph{Large Map} and \emph{Stochastic Map}. Each heatmap shows the estimates for each map position at school time. Color represents the probability of safety, ranging from certainly unsafe (dark) to certainly safe (yellow).}
    \label{fig:heatmaps}
\end{figure}

\subsection{Results}
\label{sec:exp:results}

\paragraph{(Q1) The critic accurately learns to estimate option safety in both maps.}
Figure~\ref{fig:heatmaps} shows the predictions of the pretrained neural critic for some of the options when \texttt{school-time} is active. In the \emph{Large Map}, during \texttt{school-time}, the safety concerns are avoiding the military zone and avoid driving around the school, both correctly identified by the critic which assigns high probability of a safety violations to the options moving into these areas. In the \emph{Stochastic Map}, the only safety concern is not driving towards the pedestrians at the top of the central road, again correctly identified by the critic which marks as unsafe the option moving up from the center of the map, and as safe going to the left from any position. In Appendix~\ref{app:safety-vis}, we report the complete visualizations including all options and both the \texttt{school-time} values; additionally, we report there the safety estimates obtained with a tabular critic.

\begin{figure}[t]
  \centering
  \includegraphics[width=0.8\linewidth]{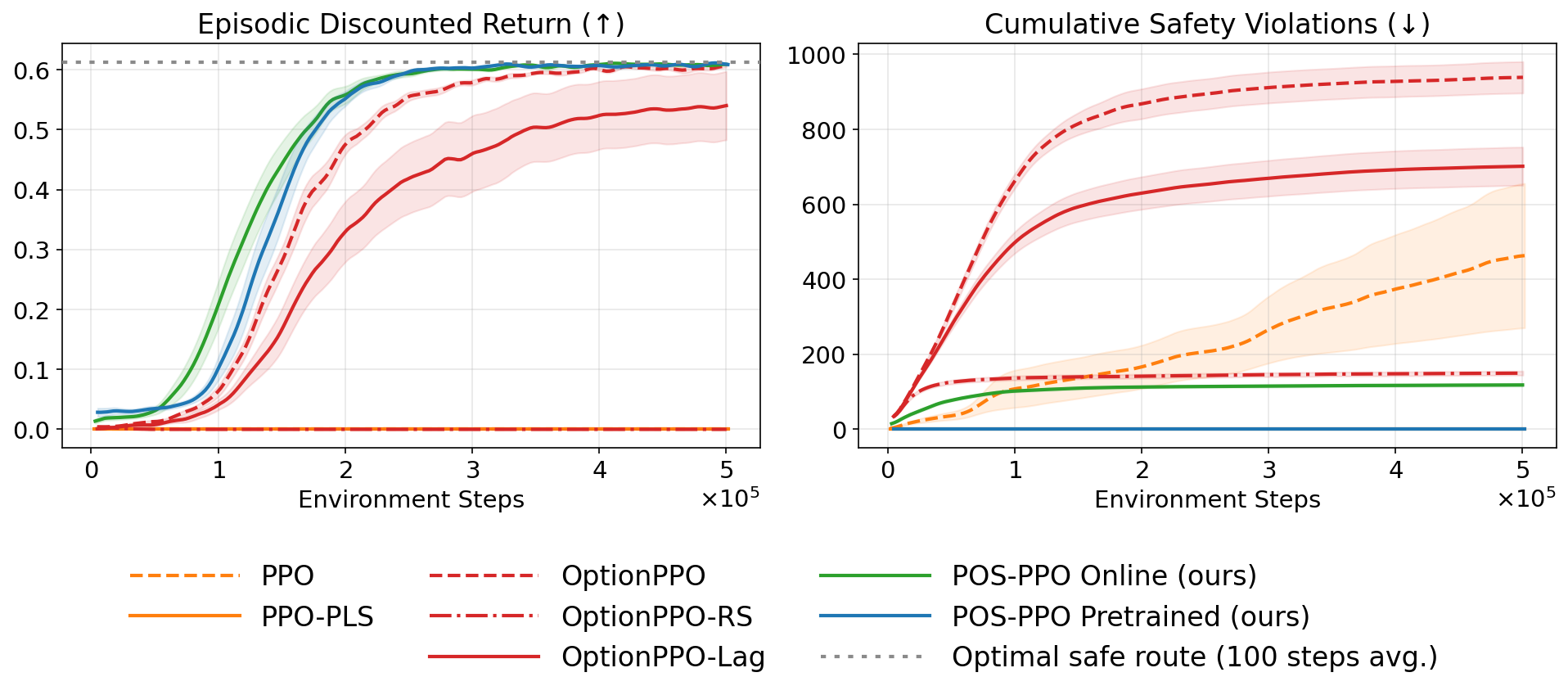}
  \caption{\textbf{Performance and safety violations on the large map.} Episodic discounted return (left) and cumulative safety violations (right) are averaged over $10$ seeds.}
  \label{fig:large-map-performance}
\end{figure}

\begin{figure}[t]
  \centering
  \includegraphics[width=0.8\linewidth]{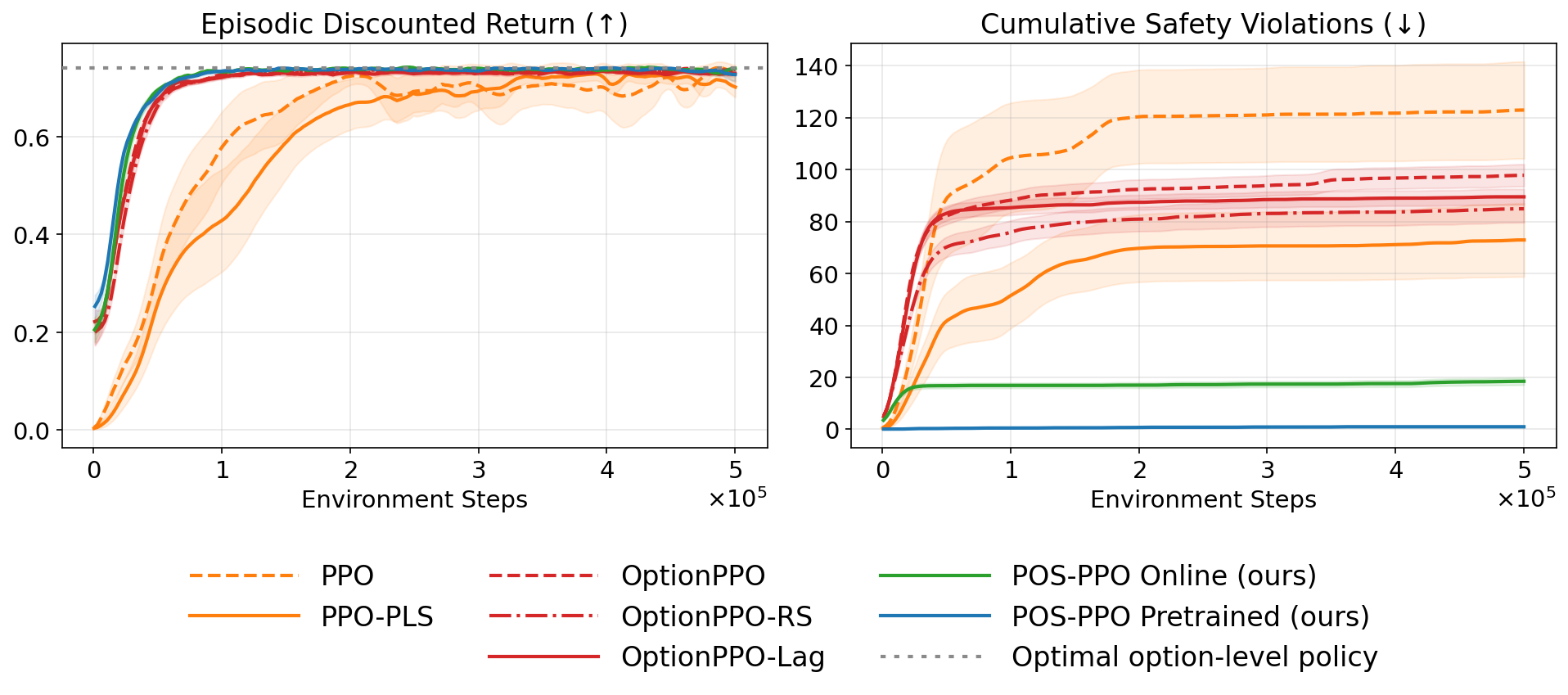}
  \caption{\textbf{Performance and safety violations on the stochastic map.} Episodic discounted return (left) and cumulative safety violations (right) are averaged over $10$ seeds.}
  \label{fig:inertia-map-performance}
\end{figure}

\paragraph{(Q2) \texttt{POS-PPO} considerably reduces safety violations compared to the baselines in both maps, while quickly learning an optimal policy.}  Figure~\ref{fig:large-map-performance} and Figure~\ref{fig:inertia-map-performance} show the episodic discounted return and cumulative safety violations for the \emph{Large Map} and \emph{Stochastic Map} experiments, respectively. In both cases, \texttt{POS-PPO} significantly outperforms the baselines in terms of safety and quickly converges to an optimal policy. In the \emph{Large Map} experiment (Figure~\ref{fig:large-map-performance}), \texttt{POS-PPO} with a pretrained critic achieves near-zero violations, while learning the critic online reduces the cumulative violations by 87\% and 83\% with respect to \texttt{OptionPPO} and \texttt{OptionPPO-Lag}, the only baselines which also consistently solve the task. The other baselines, including \texttt{PPO-PLS}, also have few violations but do not learn a policy which solves the task. 
In the \emph{Stochastic Map} experiment (Figure~\ref{fig:inertia-map-performance}), all the tested algorithms converge to an (almost) optimal policy; \texttt{POS-PPO} does so quickly, while significantly reducing violations compared to all the other baselines.  The pretrained and online versions reduce violations by 99\% and 80\% with respect to \texttt{OptionPPO-Lag}, and by 99\% and 79\% with respect to \texttt{OptionPPO-RS}.

Moreover, in the \emph{Stochastic Map}, option-level shielding proved significantly safer than its action-level counterpart. While the two are not comparable in the \emph{Large Map}, since the action-level baselines never learned to reach the goal, in this setting both of them are able to reach (near) optimal performance. However, \texttt{POS-PPO} incurs considerably fewer violations than PPO-PLS, reducing them by 98\% in the pretrained setting and 75\% in the online one. The option-level approach shows to be more effective in solving longer-horizon tasks and to be safer under stochastic dynamics.

\section{Related Work}
\label{sec:related-work}
Our work is related to several lines of research within Safe Reinforcement Learning (Safe-RL, \citep{garcia2015comprehensive}). 
\paragraph{Safe-RL via Shielding.}
Shielding approaches separate policy optimization from  constraint enforcement: a runtime mechanism evaluates the actions proposed by the policy and suppresses, replaces, or reweights actions that may violate a safety specification~\citep{konighofer2025shields}. Early work synthesized deterministic shields from temporal-logic specifications, providing formal guarantees whenever the model and specification were accurate \citep{alshiekh2018shielding}. Different variations of probabilistic shields extend this idea by comparing actions according to their probability of reaching unsafe states, e.g. using probabilistic model checking \citep{jansen2020safe}, MDP-augmentation techniques~\citep{belardinelli2025probabilistic}, or probabilistic programs~\citep{yang2023safe}. Our contribution extends the framework of~\citep{yang2023safe} to work naturally over long-horizon safety specifications, therefore inheriting their flexibility and milder MDP specification requirements compared to the other approaches. Model-predictive shielding is another related method which performs online lookahead and invokes a safe backup mechanism when a proposed action cannot be certified \citep{li2020robust,banerjee2024dynamic}. In contrast, our method learns option-level safety without the need for online planning, a complete dynamics model, or a backup controller. Shielding has also been extended to partial observability \citep{carr2023partial} and multi-agent systems \citep{elsayedaly2021multiagent}; these contributions are orthogonal to our work and could potentially be combined effectively.

\paragraph{Safe Hierarchical RL.}
Hierarchical reinforcement learning introduces temporal abstractions that divide long-horizon behavior into reusable skills or options~\citep{pateria2021hierarchical}. The Option-Critic architecture, for example, learns intra-option policies and termination conditions jointly with a policy over options \citep{bacon2017optioncritic}. Safety-aware extensions address several distinct sources of risk. Safe Option-Critic augments option learning with an uncertainty-sensitive objective that trades expected return against the variability of option outcomes \citep{jain2021safeoptioncritic}. In autonomous driving, hierarchical planning has been used to combine a learned high-level decision maker with low-level controllers designed to preserve safety \citep{li2021safehierarchical}. Other approaches formulate hierarchical decision making in terms of expected or percentile-based cost constraints \citep{lu2024constrained}. In contrast to these and other similar approaches, we assume a given option set and shield using the learned probability of safe option execution, rather than modifying low-level controllers or optimizing uncertainty-based or cost-based objectives. In this sense, our method is more similar to SQRL~\citep{srinivasan2020learning} which also learns a safety critic; however, their critic is at the action level, and their definition of safety assumes terminal unsafe events, which can be unrealistic (for example, a car bumping into another is unsafe but is not generally terminal). Moreover, our approach could be easily integrated with any method that automatically learns options.

\section{Conclusions}
\label{sec:conclusions}

We introduced a method for learning the safety of temporally extended behaviors using the option framework~\citep{sutton1999between}; we then showed how these estimates can be effectively integrated as Probabilistic Logic Shields (PLS) during training of any modern policy gradient method with control over options to yield considerably safer training.
Our results on the LogiCity simulator showed that our safety critic learns accurate safety estimates, and shielding with it considerably reduces safety violations compared to unshielded methods, standard PLS over actions, reward-shaping, and a common Lagrangian baseline, while still solving the tasks optimally. 

Finally, we outline the main limitations of our approach and related directions for future work. First, we assumed a given option set is available to the agent; an interesting extension would be to investigate how to safely learn the options themselves. Additionally, the evaluation focused on variants of the LogiCity environment; a natural next step would be to test our approach on more complex environments, especially more realistic autonomous driving benchmarks such as CARLA~\citep{dosovitskiy2017carla}.

\begin{ack}
This research has received funding from the KU Leuven
Research Funds (C14/24/092) and from the Flemish Government under the "Onderzoeksprogramma Artificiële Intelligentie (AI) Vlaanderen" programme. Y. Jiao is supported by the EU Framework Program for Research and Innovation Horizon under the Grant Agreement No 101073307 (MSCA-DN LeMuR). L. De Raedt is also supported by the European Research Council (ERC) under the European Union’s Horizon 2020 research and innovation program (Grant agreement No. 101142702) and the Wallenberg AI, Autonomous Systems and Software Program (WASP) funded by the Knut and Alice Wallenberg Foundation.
\end{ack}

\bibliographystyle{plainnat}
\bibliography{references}

\appendix

\section{Proof of Theorem~\ref{thm:fixed-point}}
\label{app:fixed-point}

We measure distances between candidate functions using the supremum norm,
$\lVert f-g\rVert_\infty
    =
    \sup_{s\in\mathcal{S}}|f(s)-g(s)|.$
Then, the following lemma bounds the difference between two candidate functions after $k$ applications of the operator $\mathcal{T}^o$ by the probability that the option remains active after $k$ steps.

\begin{lemma}[Trajectory unrolling bound]
\label{lem:trajectory-bound}
For any $k\geq 0$ and any $f,g\in[0,1]^{\mathcal{S}}$,
\[
    \left\lVert
        (\mathcal{T}^o)^k f-(\mathcal{T}^o)^k g
    \right\rVert_\infty
    \leq
    \sup_{s\in\mathcal{S}}
    \mathbf{P}(T_o>k\mid s_0=s)
    \lVert f-g\rVert_\infty.
\]
\end{lemma}

\begin{proof}
We prove by induction on $k$ that, for every $s\in\mathcal{S}$,
\begin{equation}
    \left|
        [(\mathcal{T}^o)^k f](s)
        -
        [(\mathcal{T}^o)^k g](s)
    \right|
    \leq
    \mathbf{P}(T_o>k\mid s_0=s)
    \lVert f-g\rVert_\infty.
    \label{eq:pointwise-trajectory-bound}
\end{equation}
The result then follows by taking the supremum over $s$.

For $k=0$, $(\mathcal{T}^o)^0$ is the identity and
$|f(s)-g(s)|
    \leq
    \lVert f-g\rVert_\infty.
$ Because $T_o\geq 1$, we have
$\mathbf{P}(T_o>0\mid s_0=s)=1$, so the base case holds.

Suppose that Equation~\eqref{eq:pointwise-trajectory-bound} holds at $k$ for every state. Applying $\mathcal{T}^o$ once again and subtracting causes the termination terms $\beta_o(s')$ to cancel:
\begin{align*}
&\left|
    [(\mathcal{T}^o)^{k+1}f](s)
    -
    [(\mathcal{T}^o)^{k+1}g](s)
\right|
\\
&\quad=
\left|
\mathbb{E}_{\substack{
    a\sim\pi_o(\cdot\mid s)\\[-0.5mm]
    s'\sim P(\cdot\mid s,a)
}}
\left[
    \mathbf{P}(\mathrm{safe}\mid s,a)
    \bigl(1-\beta_o(s')\bigr)
    \left(
        [(\mathcal{T}^o)^k f](s')
        -
        [(\mathcal{T}^o)^k g](s')
    \right)
\right]
\right|
\\
&\quad\leq
\mathbb{E}_{\substack{
    a\sim\pi_o(\cdot\mid s)\\[-0.5mm]
    s'\sim P(\cdot\mid s,a)
}}
\left[
    \mathbf{P}(\mathrm{safe}\mid s,a)
    \bigl(1-\beta_o(s')\bigr)
    \left|
        [(\mathcal{T}^o)^k f](s')
        -
        [(\mathcal{T}^o)^k g](s')
    \right|
\right]
\\
&\quad\leq
\mathbb{E}_{\substack{
    a\sim\pi_o(\cdot\mid s)\\[-0.5mm]
    s'\sim P(\cdot\mid s,a)
}}
\left[
    \mathbf{P}(\mathrm{safe}\mid s,a)
    \bigl(1-\beta_o(s')\bigr)
    \mathbf{P}(T_o>k\mid s_0=s')
\right]
\lVert f-g\rVert_\infty
\\
&\quad\leq
\mathbb{E}_{\substack{
    a\sim\pi_o(\cdot\mid s)\\[-0.5mm]
    s'\sim P(\cdot\mid s,a)
}}
\left[
    \bigl(1-\beta_o(s')\bigr)
    \mathbf{P}(T_o>k\mid s_0=s')
\right]
\lVert f-g\rVert_\infty
\\
&\quad=
\mathbf{P}(T_o>k+1\mid s_0=s)
\lVert f-g\rVert_\infty.
\end{align*}
The first inequality follows from
$|\mathbb{E}[X]|\leq\mathbb{E}[|X|]$.
The second applies the inductive hypothesis, and the third follows since
$\mathbf{P}(\mathrm{safe}\mid s,a)\leq 1$.
The last equality holds 
since the option remains active for more than $k+1$ steps precisely when it does not terminate after the first transition and, from $s'$, its remaining execution lasts for more than $k$ steps. This completes the induction.
\end{proof}

\begin{proof}[Proof of Theorem~\ref{thm:fixed-point}]
We now need to prove that repeated applications of the operator $\mathcal{T}^o$ reach a unique fixed point.  \\\\
\emph{Range preservation.}
We first verify that $\mathcal{T}^o$ maps
$[0,1]^{\mathcal{S}}$ into itself. For any
$f\in[0,1]^{\mathcal{S}}$,
$\beta_o(s')
    +
    \bigl(1-\beta_o(s')\bigr)f(s')
$
is a convex combination of $1$ and $f(s')$ and therefore lies in $[0,1]$. Multiplying it by
$\mathbf{P}(\mathrm{safe}\mid s,a)\in[0,1]$ preserves this range, as does taking the expectation over $a$ and $s'$. Hence,
$\mathcal{T}^o f\in[0,1]^{\mathcal{S}}$.

By Assumption~\ref{ass:bounded},
$
    \mathbf{P}(T_o>N\mid s_0=s)=0
$
for every $s\in\mathcal{S}$. Setting $k=N$ in Lemma~\ref{lem:trajectory-bound} therefore gives
$    \left\lVert
        (\mathcal{T}^o)^N f-(\mathcal{T}^o)^N g
    \right\rVert_\infty
    =
    0
$
for every $f,g\in[0,1]^{\mathcal{S}}$. Thus, $(\mathcal{T}^o)^N$ is a constant map: there exists some
$c\in[0,1]^{\mathcal{S}}$ such that
$
    (\mathcal{T}^o)^N f=c
$
for every $f\in[0,1]^{\mathcal{S}}$.

\emph{Existence.}
In particular, $(\mathcal{T}^o)^N c=c$. Then, applying $\mathcal{T}^o$ gives
$(\mathcal{T}^o)^{N+1}c=\mathcal{T}^o c.
$
On the other hand, range preservation implies
$\mathcal{T}^o c\in[0,1]^{\mathcal{S}}$, and 
$(\mathcal{T}^o)^N$ being a constant map gives
$(\mathcal{T}^o)^{N+1}c
    =
    (\mathcal{T}^o)^N(\mathcal{T}^o c)
    =
    c.
$
Therefore, $\mathcal{T}^o c=c$, so $c$ is a fixed point.

\emph{Uniqueness.}
Suppose that $f^\star$ is another fixed point of $\mathcal{T}^o$. Then
$
(\mathcal{T}^o)^N f^\star=f^\star.
$
Because $(\mathcal{T}^o)^N$ maps every element of
$[0,1]^{\mathcal{S}}$ to $c$, it follows that $f^\star=c$. Hence, the fixed point is unique.

\emph{Convergence.}
For any initial $f_0\in[0,1]^{\mathcal{S}}$,
$
(\mathcal{T}^o)^N f_0=c.
$
Because $c$ is a fixed point, every subsequent iterate also equals $c$. Thus, iteratively applying of $\mathcal{T}^o$ reaches the unique fixed point after at most $N$ applications.
\end{proof}

\section{Experimental details}
\label{app:hyper}

\subsection{Environment}
\label{app:env}

\begin{table}[t]
\centering
\caption{\textbf{Environment parameters.} Positions are given as (row, column) on the grid. All other environment settings are identical between the two maps.}
\label{tab:exp-env}
\begin{tabular}{lcc}
\toprule
 & \emph{Large Map} & \emph{Stochastic Map} \\
\midrule
Grid size            & $43 \times 43$        & $23 \times 23$ \\
Start cell           & $(30, 32)$            & $(20, 12)$ \\
Goal cell            & $(0, 42)$             & $(0, 12)$ \\
Step limit           & $800$                 & $250$ \\
Pedestrians          & $8$                   & $3$ \\
Military zone        & rows $14$--$17$, cols $14$--$42$ & none \\
Inertia probability  & none                  & $0.3$ \\
Discount $\gamma$    & $0.995$               & $0.99$ \\
\bottomrule
\end{tabular}
\end{table}

The environment used for the experiments is implemented in LogiCity \citep{li2024logicity} (MIT License). 
\autoref{tab:exp-env} collects the parameters that differ between the two maps we tested.

\begin{figure}[t]
\centering
\begin{subfigure}[t]{0.2\linewidth}
  \centering\includegraphics[width=\linewidth]{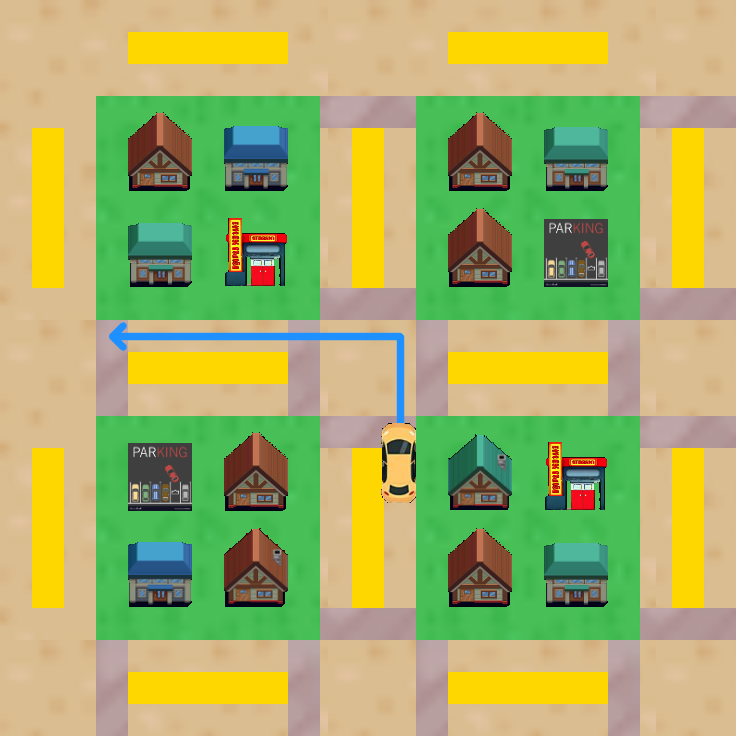}
  \caption{\emph{left}}
\end{subfigure}\hfill
\begin{subfigure}[t]{0.2\linewidth}
  \centering\includegraphics[width=\linewidth]{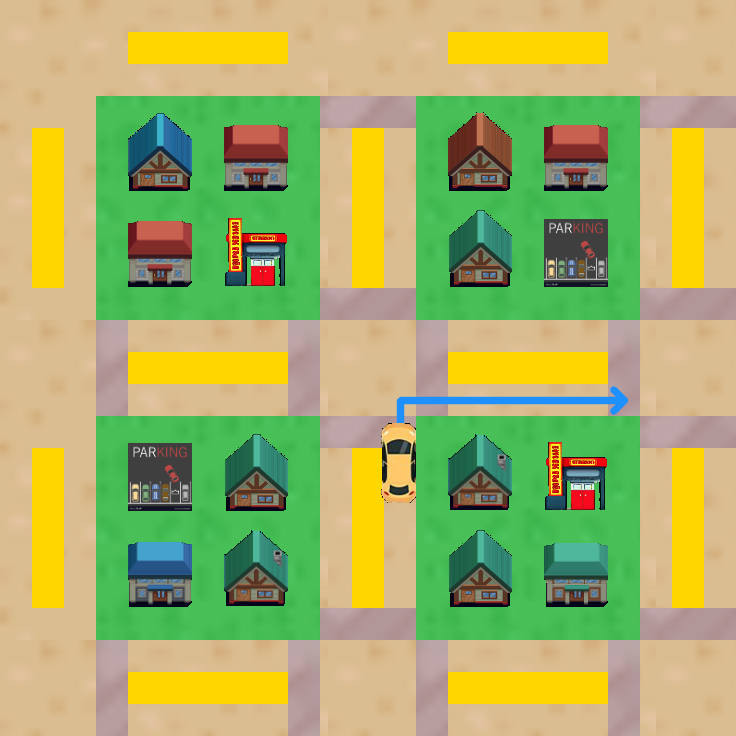}
  \caption{\emph{right}}
\end{subfigure}\hfill
\begin{subfigure}[t]{0.2\linewidth}
  \centering\includegraphics[width=\linewidth]{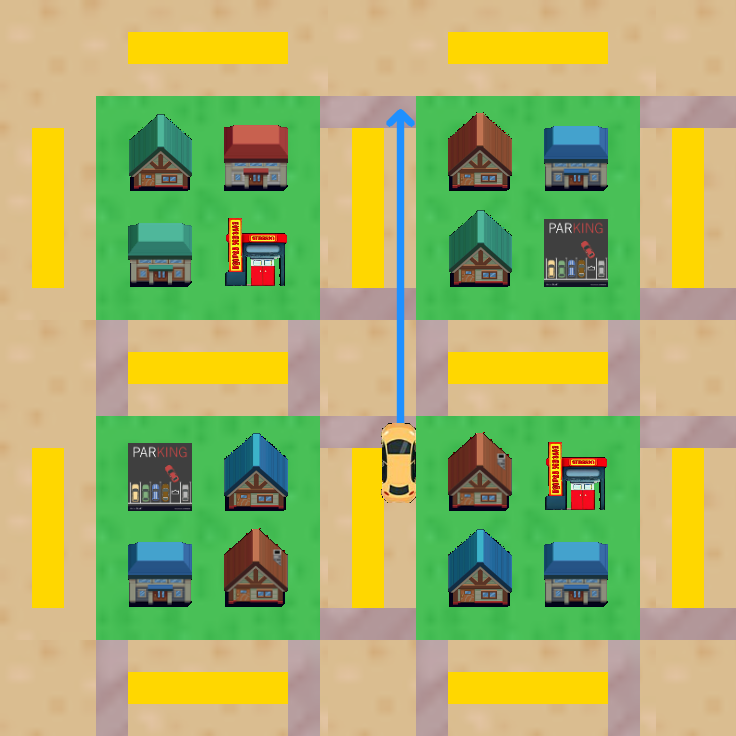}
  \caption{\emph{up}}
\end{subfigure}\hfill
\begin{subfigure}[t]{0.2\linewidth}
  \centering\includegraphics[width=\linewidth]{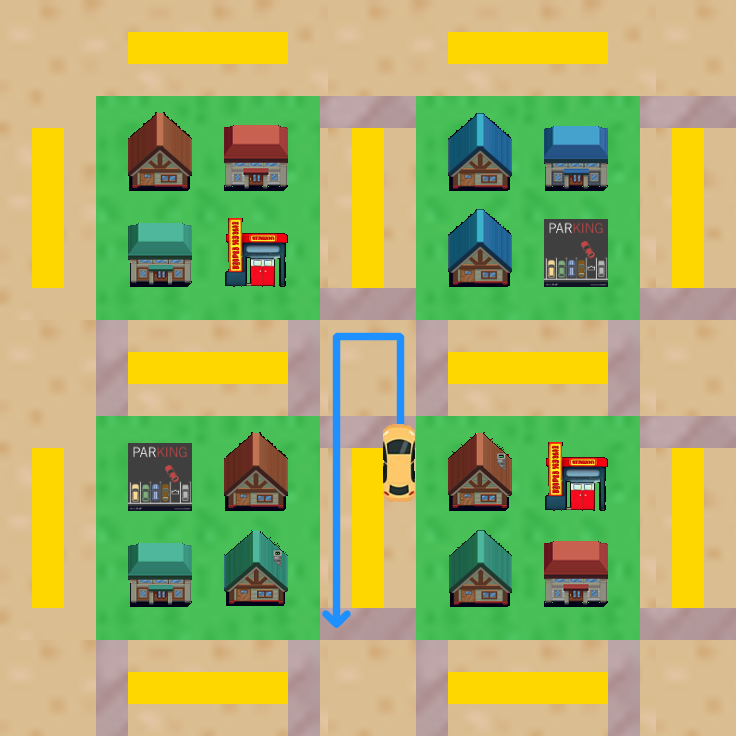}
  \caption{\emph{down}}
\end{subfigure}
\caption{\textbf{The four options.} Each panel traces the trajectory followed from a common starting cell, drawn schematically on a small grid.}
\label{fig:options}
\end{figure}

\subsection{Options}
The options are assumed to be given in this work. \autoref{fig:options} shows the available ones: each one takes the agent from its current position to the next intersection in one of the four cardinal directions. They can be selected in any state, and terminate once they reach the target intersection, or immediately if no further intersection in the chosen direction is present.

\begin{lstlisting}[style=problog, float=t, label=lst:shield, caption={\textbf{Safety specification for the \emph{Large Map}.}}]
action(0)::action(left).
action(1)::action(right).
action(2)::action(up).
action(3)::action(down).
action(4)::action(stop).

% Other-agent sensors (indices 0-3)
sensor_value(0)::other_agent(left).
sensor_value(1)::other_agent(right).
sensor_value(2)::other_agent(up).
sensor_value(3)::other_agent(down).

% Military-area sensors (indices 4-7)
sensor_value(4)::military_area(left).
sensor_value(5)::military_area(right).
sensor_value(6)::military_area(up).
sensor_value(7)::military_area(down).

transition(left,left).
transition(right,right).
transition(up,up).
transition(down,down).

% Unsafe if the next cell is occupied by another agent
unsafe_next :- action(A), transition(A,NextPos), other_agent(NextPos).
% Unsafe if the next cell falls in the military zone
unsafe_next :- action(A), transition(A,NextPos), military_area(NextPos).

safe_next :- action(stop).
safe_next :- \+ unsafe_next.
\end{lstlisting}

\subsection{Safety specification}
\label{app:spec}

\autoref{lst:shield} gives the ProbLog program used on the \emph{Large Map}. It maps four adjacent-cell pedestrian sensors, four military-zone sensors and the selected action to $\mathbf{P}(\text{safe}\mid s,a)$. The program reads as follows: the first block shows for each of the five actions the probability the policy assigns to it, and the next two blocks show the corresponding sensor readings in each of the four directions for the pedestrian and military-zone atoms; the \texttt{transition} facts map an action to the cell it moves into. The two \texttt{unsafe\_next} rules then state that an action is unsafe if it brings the agent into a cell holding a pedestrian or belonging to the military zone. Finally, \emph{stop} is safe by construction and any other action is safe when it is not unsafe. Querying \texttt{safe\_next} yields $\mathbf{P}(\text{safe}\mid s,a)$. 

The \emph{Stochastic Map} uses the same program with the four \texttt{military\_area} sensors and the second \texttt{unsafe\_next} clause removed, since that map has no military zone. 

\begin{table}[t]
\centering
\caption{\textbf{PPO hyperparameters.} \emph{Opt} and \emph{act} denote the option-level and action-level arms. Rollout length is measured in buffer entries, which are options at the option level and primitive actions at the action level.}
\label{tab:exp-ppo}
\begin{tabular}{lcccc}
\toprule
 & \multicolumn{2}{c}{\emph{Large Map}} & \multicolumn{2}{c}{\emph{Stochastic Map}} \\
\cmidrule(lr){2-3}\cmidrule(lr){4-5}
Hyperparameter & Opt & Act & Opt & Act \\
\midrule
Learning rate        & $10^{-4}$ & $10^{-4}$ & $10^{-4}$ & $10^{-4}$ \\
Discount $\gamma$    & $0.995$   & $0.995$   & $0.99$    & $0.99$ \\
Clip range           & $0.15$    & $0.15$    & $0.15$    & $0.15$ \\
Value-function coef. & $0.5$     & $0.5$     & $0.5$     & $0.5$ \\
Entropy coef.        & $0.01$    & $0.01$    & $0.01$    & $0.01$ \\
Epochs per update    & $8$       & $8$       & $8$       & $8$ \\
Rollout length       & $512$     & $2048$    & $128$     & $512$ \\
Minibatch size       & $128$     & $512$     & $32$      & $128$ \\
\bottomrule
\end{tabular}
\end{table}
\subsection{PPO hyperparameters}
\label{app:ppo}

Every presented method trains a PPO policy and they all use the configuration in \autoref{tab:exp-ppo}. Most hyperparameters are shared across all methods, although between action-level and option-level baselines the rollout length and the minibatch size differ. These are chosen so that the two levels consume comparable numbers of environment steps per update.

\begin{table}[t]
\centering
\caption{\textbf{Neural safety-critic settings.}}
\label{tab:exp-critic}
\begin{tabular}{lcc}
\toprule
 & \emph{Large Map} & \emph{Stochastic Map} \\
\midrule
Hidden layers                & $64,64$            & $32$               \\
Output power $k$             & $4$                & $2$                \\
Learning rate                & $5\times10^{-4}$   & $2\times10^{-3}$   \\
Gradient steps per decision  & $2$                & $8$                \\
Replay-buffer size           & $5\times10^{5}$    & $5\times10^{5}$    \\
Minibatch size               & $128$              & $128$              \\
Optimizer                    & Adam                & Adam                \\
Polyak coefficient $\tau$    & $0.005$            & $0.005$            \\
Online update frequency      & Every option decision & Every option decision \\
\bottomrule
\end{tabular}
\end{table}

\subsection{Safety critic}
\label{app:critic}

\paragraph{Tabular critic.} The tabular critic stores one entry per $(\mathit{row}, \mathit{col}, \texttt{school-time}, o)$ tuple, initialised at $0.5$ everywhere in both training regimes. In the pretrained regime the step size starts at $\alpha = 0.1$ and is halved whenever the updates plateau, with training stopped once the entries stabilise across a full pass. In the online regime the step size is fixed at $\alpha = 1$ on the \emph{Stochastic Map} and reduced to $\alpha = 0.99$ on the \emph{Large Map}. 

\paragraph{Neural critic.}
The neural critic is an MLP that takes the three-dimensional observation as input and produces one safety estimate per option. Because cumulative option-safety probabilities can be very small, we parameterize each output as $
\hat{P}(\mathrm{safe}\mid s,o)=\sigma(z_o)^k,
$ where $k$ is a fixed positive integer. This transformation changes the parameterization of the critic but not the quantity it estimates, since the powered output is trained directly against the TD target in probability space. It allows small probabilities to be represented with less extreme logits: for example, an output of $10^{-6}$ requires $z_o\approx-13.8$ when $k=1$, but only $z_o\approx-3.4$ when $k=4$. We minimize the mean squared TD error using Adam~\citep{kingma2014adam} with minibatches of size $128$. A target network, updated by Polyak averaging with $\tau=0.005$, provides the bootstrap estimates. \autoref{tab:exp-critic} gives the per-map settings.

\subsection{Baselines and hyperparameter selection}
\label{app:baselines}

\texttt{PPO} and \texttt{PPO-PLS} operate on low-level actions, the latter applying the one-step shield of \citet{yang2023safe} to the program of \autoref{app:spec}. \texttt{OptionPPO} is the same PPO applied to the policy over options with no safety mechanism. For these baselines, the only hyperparameters are the standard PPO ones reported in Table~\ref{tab:exp-ppo}.
\begin{table}[t]
\centering
\caption{\textbf{Reward-shaping penalty sweep.} Episodic discounted return and cumulative violations at the end of training, mean $\pm$ SEM over five seeds per setting. Selected values in bold}
\label{tab:exp-rs-sweep}
\begin{tabular}{lcccc}
\toprule
 & \multicolumn{2}{c}{\emph{Large Map}} & \multicolumn{2}{c}{\emph{Stochastic Map}} \\
\cmidrule(lr){2-3}\cmidrule(lr){4-5}
Penalty & Return & Violations & Return & Violations \\
\midrule
$-0.1$  & $0.000 \pm 0.000$ & $139.9 \pm 2.6$  & $0.716 \pm 0.009$ & $112.0 \pm 2.3$ \\
$-0.25$ & $0.000 \pm 0.000$ & $144.4 \pm 5.8$  & $0.693 \pm 0.016$ & $110.7 \pm 4.7$ \\
$-0.5$  & $0.000 \pm 0.000$ & $146.9 \pm 2.7$  & $0.681 \pm 0.016$ & $110.0 \pm 8.0$ \\
$-0.75$ & $\mathbf{0.000 \pm 0.000}$ & $\mathbf{139.4 \pm 2.4}$ & $0.704 \pm 0.007$ & $104.3 \pm 2.6$ \\
$-0.9$  & $0.000 \pm 0.000$ & $146.7 \pm 1.8$  & $\mathbf{0.702 \pm 0.004}$ & $\mathbf{94.1 \pm 4.9}$ \\
$-1.0$  & $0.000 \pm 0.000$ & $144.5 \pm 10.4$ & $0.679 \pm 0.015$ & $98.6 \pm 9.2$ \\
\bottomrule
\end{tabular}
\end{table}

\paragraph{Reward shaping.} \texttt{OptionPPO-RS} adds a constant negative reward for each violated task rule. The penalty was swept over $-0.1$, $-0.25$, $-0.5$, $-0.75$, $-0.9$ and $-1.0$ with the PPO configuration held fixed; \autoref{tab:exp-rs-sweep} reports the outcome. For both maps the penalty that held the least violations was selected.

\paragraph{Lagrangian.} \texttt{OptionPPO-Lag} treats the safety specification as in a constrained MDP, considering a cost of $1$ for each safety violation. The sweep values for the cost limit $d$ and the multiplier step size are reported in \autoref{tab:exp-lag-sweep}. For both maps the selected hyperparameters are the ones yielding the least violations among the ones solving the task.

\begin{table}[t]
\centering
\caption{\textbf{Lagrangian sweep.} Each cell gives episodic discounted return (top) and cumulative safety violations (bottom) at the end of training, as mean $\pm$ standard error over three seeds.
}
\label{tab:exp-lag-sweep}
\small
\providecommand{\lc}[4]{\shortstack{$#1 \pm #2$\\[1pt] $#3 \pm #4$}}
\providecommand{\lcb}[4]{\shortstack{$\mathbf{#1} \pm \mathbf{#2}$\\[1pt] $\mathbf{#3} \pm \mathbf{#4}$}}
\setlength{\tabcolsep}{4pt}
\begin{tabular}{lcccc}
\toprule
 & \multicolumn{4}{c}{Multiplier step size} \\
\cmidrule(lr){2-5}
Cost limit $d$ & $0.01$ & $0.02$ & $0.05$ & $0.15$ \\
\midrule
\multicolumn{5}{l}{\emph{Large Map}} \\
$0$    & \lc{0.607}{0.002}{817}{57.0} & \lc{0.405}{0.165}{638}{63.3} & \lc{0.203}{0.165}{370}{97.2} & \lc{0.000}{0.000}{206}{17.2} \\
\addlinespace
$0.05$ & \lc{0.608}{0.002}{828}{39.1} & \lcb{0.602}{0.004}{773}{32.4} & \lc{0.094}{0.076}{339}{54.5} & \lc{0.000}{0.000}{211}{16.4} \\
\addlinespace
$0.15$ & \lc{0.609}{0.003}{864}{11.3} & \lc{0.408}{0.167}{593}{76.3} & \lc{0.101}{0.083}{329}{36.9} & \lc{0.000}{0.000}{270}{11.0} \\
\addlinespace
$0.3$  & \lc{0.610}{0.003}{835}{5.3} & \lc{0.606}{0.002}{832}{49.9} & \lc{0.402}{0.164}{481}{97.9} & \lc{0.164}{0.068}{350}{43.9} \\
\midrule
\multicolumn{5}{l}{\emph{Stochastic Map}} \\
$0$    & \lc{0.741}{0.003}{88}{4.5} & \lc{0.734}{0.006}{90}{9.2} & \lc{0.731}{0.008}{95}{2.3} & \lc{0.703}{0.019}{86}{7.7} \\
\addlinespace
$0.02$ & \lc{0.735}{0.006}{93}{6.2} & \lc{0.735}{0.007}{91}{2.7} & \lcb{0.717}{0.004}{77}{2.9} & \lc{0.739}{0.004}{93}{2.2} \\
\addlinespace
$0.05$ & \lc{0.735}{0.003}{92}{0.5} & \lc{0.739}{0.005}{87}{3.1} & \lc{0.733}{0.004}{83}{2.1} & \lc{0.738}{0.006}{82}{5.0} \\
\addlinespace
$0.15$ & \lc{0.733}{0.008}{95}{2.4} & \lc{0.745}{0.005}{95}{5.8} & \lc{0.745}{0.001}{80}{3.6} & \lc{0.736}{0.004}{86}{2.8} \\
\bottomrule
\end{tabular}
\end{table}

\subsection{Evaluation protocol}
\label{app:eval}

Every experiment is run for $5 \times 10^{5}$ environment steps with ten seeds, and all curves show the mean across seeds with the standard error of the mean. Episodic discounted return is accumulated over each episode, so an episode reaching the goal after $L$ steps scores exactly $\gamma^{L-1}$. Cumulative safety violations are the running total of violated safety specifications over the whole run.

\subsection{Compute}
\label{app:compute}

All experiments were run on a single Apple M1 Pro laptop with $10$ CPU cores and $16$ GB of memory, under macOS $15.7$, with Python $3.11$, PyTorch $2.9$ and Stable-Baselines3 $2.7$. \autoref{tab:exp-compute} reports the time (in seconds) needed for each method to reach $100k$ environment steps.
The table shows that all methods have comparable training times.

\begin{table}[t]
\centering
\caption{\textbf{Training time per method.} \emph{Time} is wall-clock seconds per $10^5$ environment steps, the median over three sequential runs.}
\label{tab:exp-compute}
\small
\begin{tabular}{lrr}
\toprule
Method & \emph{Large} & \emph{Stochastic} \\
\midrule
\texttt{PPO}                 & $99.6$  & $105.9$ \\
\texttt{PPO-PLS}             & $181.7$ & $126.2$ \\
\texttt{OptionPPO}           & $114.8$ & $109.1$ \\
\texttt{OptionPPO-RS}        & $95.2$  & $108.3$ \\
\texttt{OptionPPO-Lag}       & $114.9$ & $111.8$ \\
\texttt{POS-PPO-P} (tabular) & $100.3$ & $110.6$ \\
\texttt{POS-PPO-O} (tabular) & $143.1$ & $137.7$ \\
\texttt{POS-PPO-P} (neural)  & $103.7$ & $110.8$ \\
\texttt{POS-PPO-O} (neural)  & $155.7$ & $193.2$ \\
\bottomrule
\end{tabular}
\end{table}

\section{Additional safety visualizations}
\label{app:safety-vis}

\begin{figure}[!ht]
  \centering
  \includegraphics[width=0.8\linewidth]{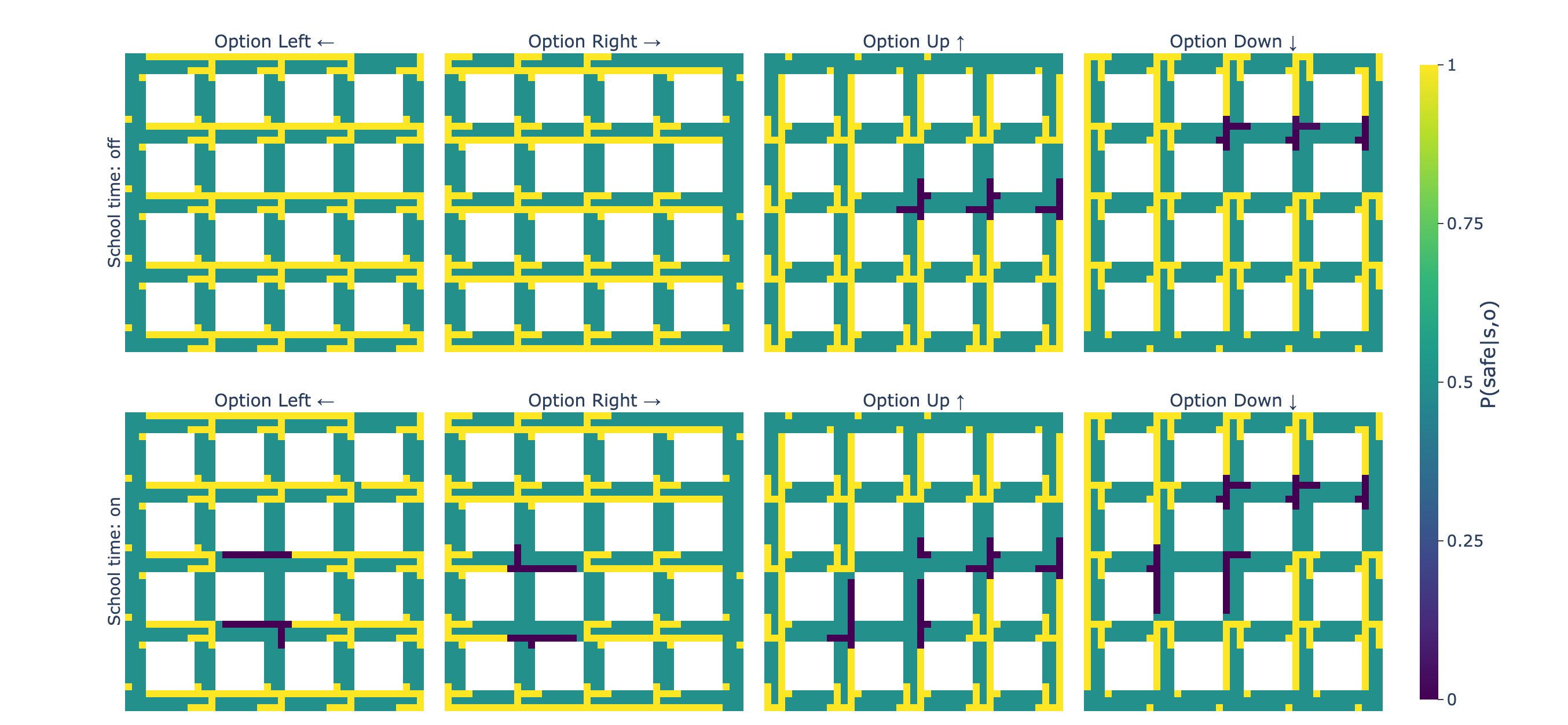}
  \caption{\textbf{Pretrained Tabular Critic, Large Map.}  The tabular critic estimates the safety of each option in each state. The bottom row is for school time, the top one for no school time. The representation matches \autoref{fig:large-map-heatmap}}
  \label{fig:large-map-tabular-heatmap-complete}
\end{figure}

\begin{figure}[t]
  \centering
  \includegraphics[width=0.8\linewidth]{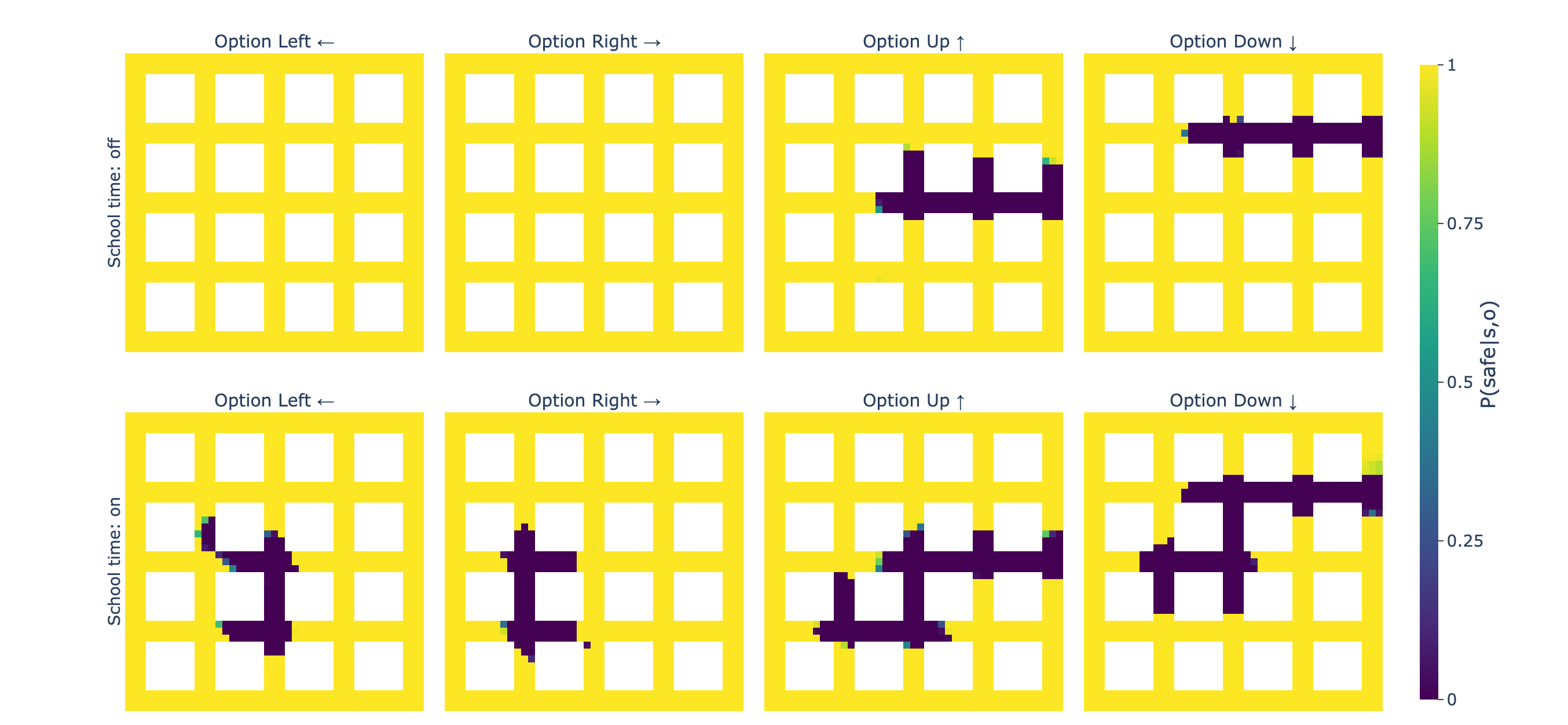}
  \caption{\textbf{Pretrained Neural Critic, Large Map.}  The neural critic estimates the safety of each option in each state. The bottom row is for school time, the top one for no school time. The representation matches \autoref{fig:large-map-heatmap}}
  \label{fig:large-map-neural-heatmap-complete}
\end{figure}

\begin{figure}[t]
  \centering
  \includegraphics[width=0.8\linewidth]{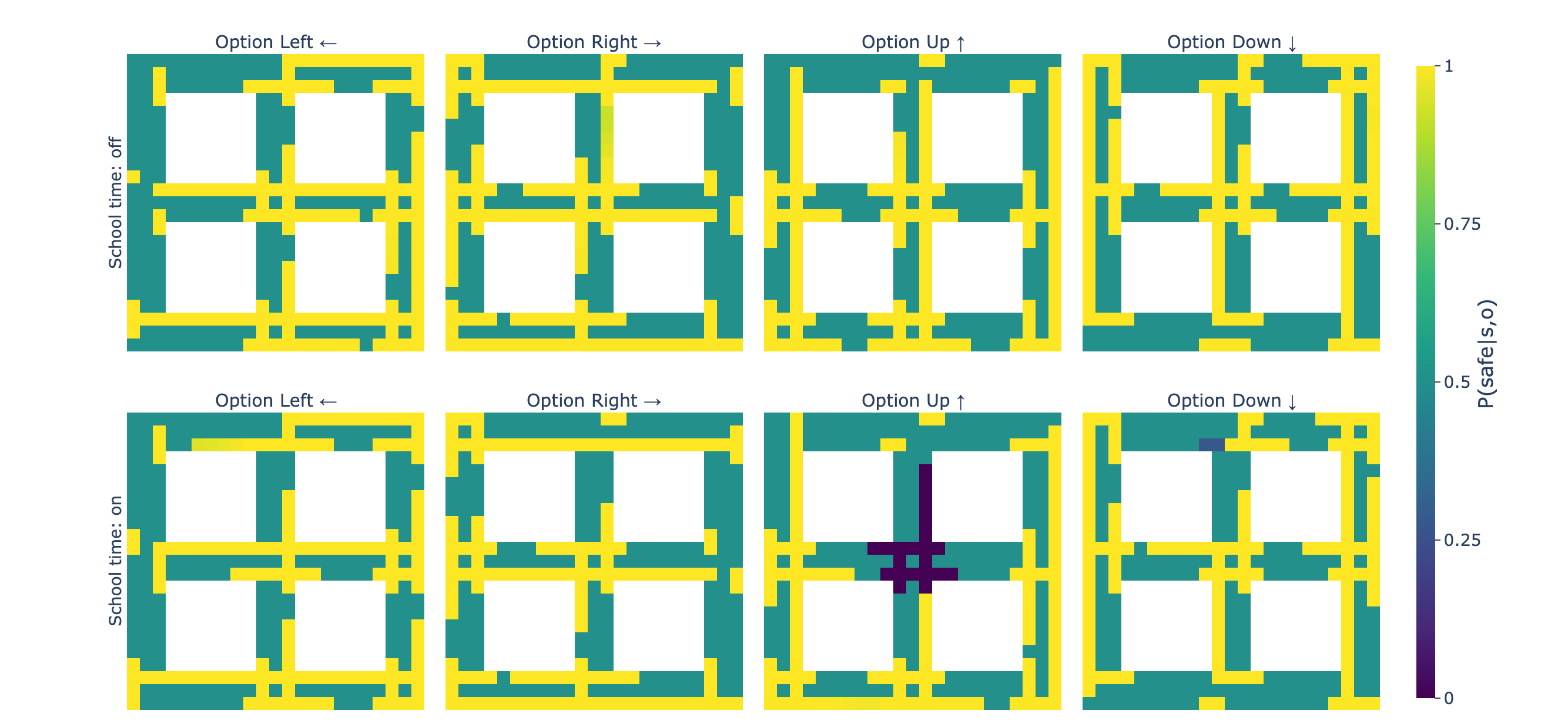}
  \caption{\textbf{Pretrained Tabular Critic, Stochastic Map.}  The tabular critic estimates the safety of each option in each state. The bottom row is for school time, the top one for no school time. The representation matches \autoref{fig:inertia-map-heatmap}}
  \label{fig:inertia-map-tabular-heatmap-complete}
\end{figure}

\begin{figure}[t]
  \centering
  \includegraphics[width=0.8\linewidth]{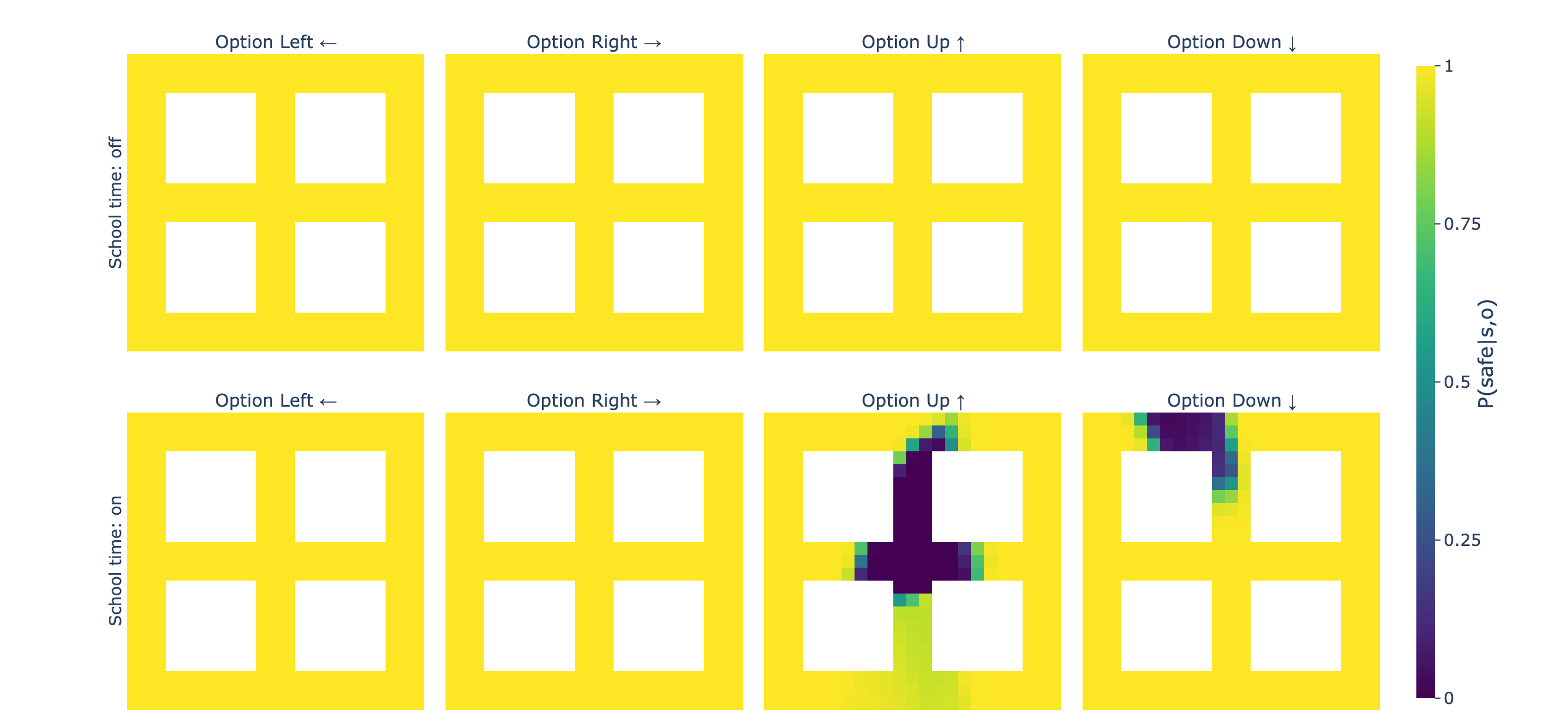}
  \caption{\textbf{Pretrained Neural Critic, Stochastic Map.}  The neural critic estimates the safety of each option in each state. The bottom row is for school time, the top one for no school time. The representation matches \autoref{fig:inertia-map-heatmap}}
  \label{fig:inertia-map-neural-heatmap-complete}
\end{figure}

The figures below show the full visualizations of the learned critics in both tabular and neural instantiation. \autoref{fig:large-map-tabular-heatmap-complete} and \autoref{fig:large-map-neural-heatmap-complete} for the \emph{Large Map}, and \autoref{fig:inertia-map-tabular-heatmap-complete} and \autoref{fig:inertia-map-neural-heatmap-complete} for the \emph{Inertia Map}.

\section{Tabular vs Neural Critic Comparison}
\label{app:neur-vs-tab}

\begin{figure}[!ht]
  \centering
  \includegraphics[width=1\linewidth]{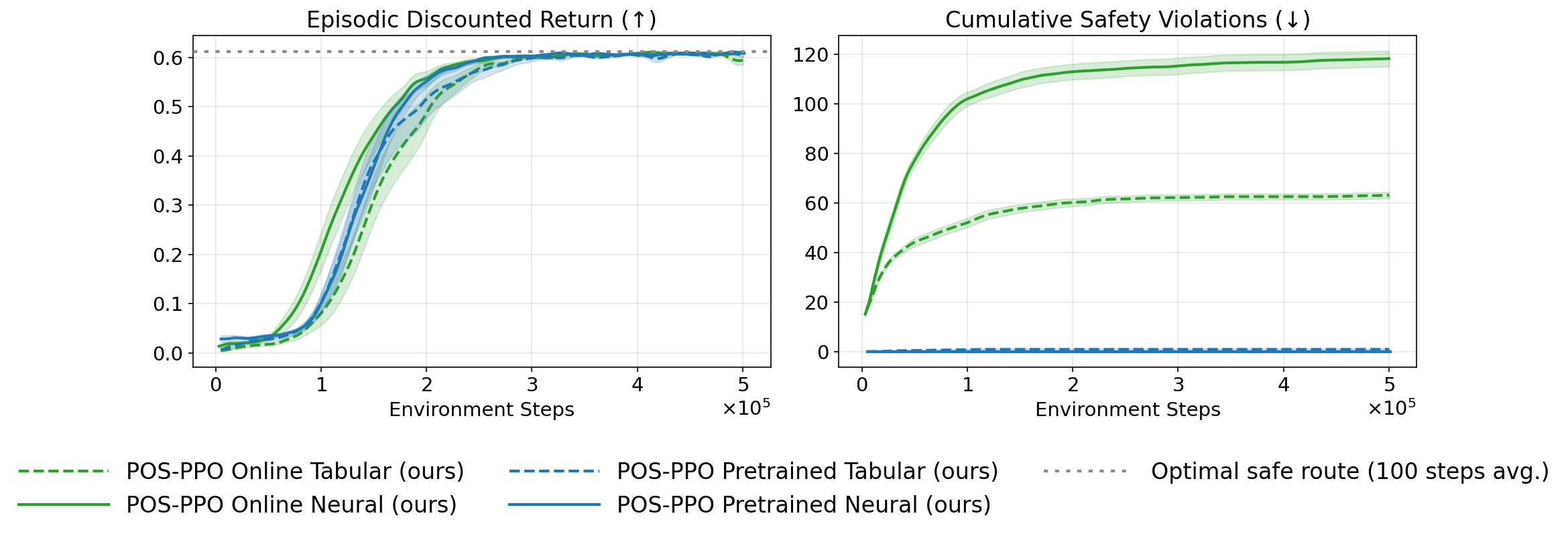}
  \caption{\textbf{Performance and Safety Violations Large Map, Tabular vs Neural Critic.}  Episodic discounted return (left) and cumulative safety violations (right) averaged over 10 seeds.}
  \label{fig:large-map-performance-tvn}
\end{figure}

\begin{figure}[!ht]
  \centering
  \includegraphics[width=1\linewidth]{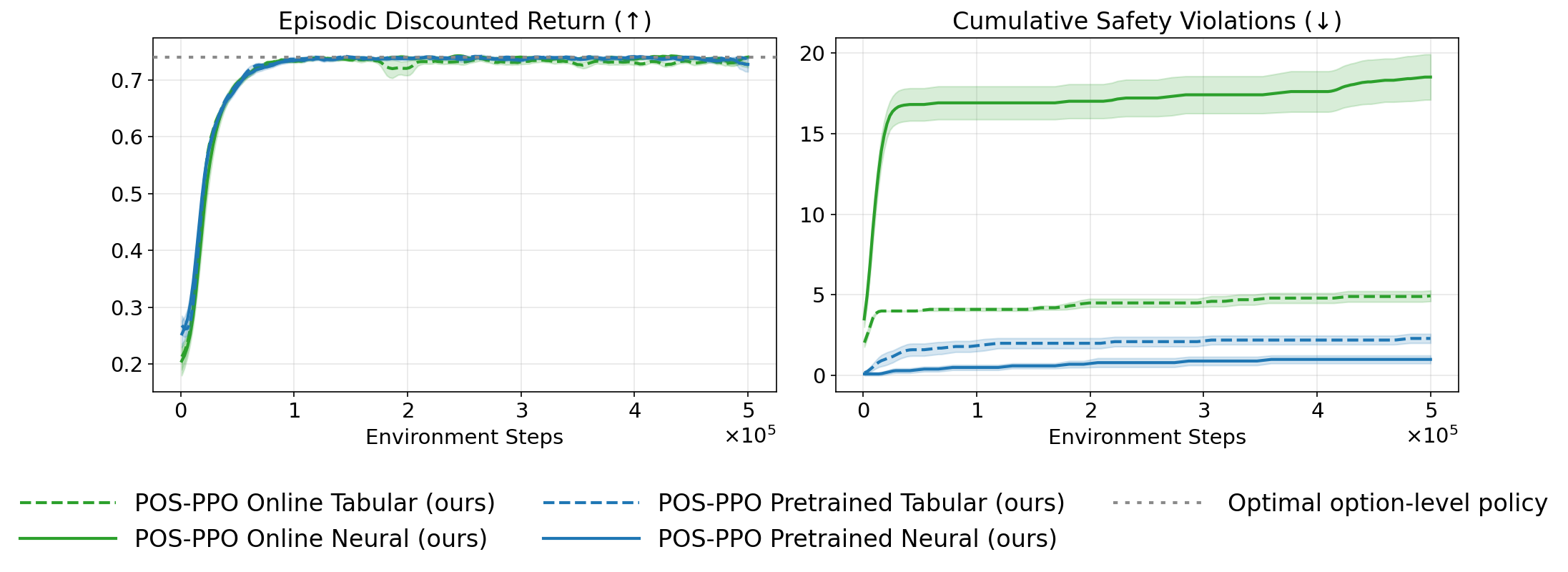}
  \caption{\textbf{Performance and Safety Violations Large Map, Tabular vs Neural Critic.}  Episodic discounted return (left) and cumulative safety violations (right) averaged over 10 seeds.}
  \label{fig:inertia-map-performance-tvn}
\end{figure}

\autoref{fig:large-map-performance-tvn} and \autoref{fig:inertia-map-performance-tvn} show the differences in performance and safety violations of the tabular and neural instantiations of the critic.

\end{document}